\documentclass{article}

\usepackage{arxiv}

\usepackage[utf8]{inputenc} % allow utf-8 input
\usepackage[T1]{fontenc}    % use 8-bit T1 fonts
\usepackage{hyperref}       % hyperlinks
\usepackage{url}            % simple URL typesetting
\usepackage{booktabs}       % professional-quality tables
\usepackage{amsfonts}       % blackboard math symbols
\usepackage{nicefrac}       % compact symbols for 1/2, etc.
\usepackage{microtype}      % microtypography
\usepackage{lipsum}

\usepackage{url}
\usepackage{mathtools}
\usepackage{amsmath} 
\usepackage{subfigure}
\allowdisplaybreaks
\usepackage{amssymb}
\usepackage{comment} 
\usepackage{dsfont}
\usepackage{titling}
\newcommand{\eq}[1]{{Eq~(#1)}}

\usepackage{ulem}
\usepackage{amsthm}
\usepackage{authblk}
\usepackage{algorithm}% http://ctan.org/pkg/algorithms
\usepackage{algpseudocode}% http://ctan.org/pkg/algorithmicx 

\newcommand{\cA}{{\mathcal{A}}}
\newcommand{\cT}{{\mathcal{T}}}
\newcommand{\cM}{{\mathcal{M}}}

\newcommand{\cS}{{\mathcal{S}}} 
\newcommand{\KL}{{\mathbf{KL}}} 
\newcommand{\cR}{{\mathcal{R}}} 
\newcommand{\RB}{{\cR_{\text{B}}}}

\newcommand{\cE}{{\mathds{E}}}

\newcommand{\RE}{{\cR_{{T}}}}
 
\newcommand{\dE}{{d_{{T}}}}
\newcommand{\pE}{{\pi_{{T}}}} 
\newcommand{\dB}{{d_{B}}}
\newcommand{\dP}{{d_\pi}}

\newcommand{\dMIX}{{d_{\text{mix}}}}

\newcommand{\SAIL}{{\textit{SAIL~}}}
\newcommand{\IL}{{\textit{IL~}}}

\newcommand{\GAIL}{{\textit{GAIL~}}}

\newcommand{\st}{{\textrm{s.t.}}}
\newcommand{\ie}{{\textrm{i.e.}}}

\newtheorem{assumption}{Assumption}
\newtheorem{remark}{Remark}

\author[1]{Zhuangdi Zhu}
\author[1]{Kaixiang Lin}
\author[2]{Bo Dai}
\author[1]{Jiayu Zhou}  
\affil[1]{Michigan State University}
\affil[2]{Google Research}
\affil[1]{\textit{zhuzhuan,linkaixi,jiayuz}@msu.edu}
\affil[2]{\textit{bodai}@google.com}

\title{Learning Sparse Rewarded Tasks from Sub-Optimal Demonstrations}
\date{}
 
\begin{document} 
\maketitle

\begin{abstract} 
    Model-free deep reinforcement learning (RL) has demonstrated its superiority
    on many complex sequential decision-making problems. However, heavy dependence on dense rewards and high sample-complexity impedes the wide adoption of these methods in real-world scenarios. On the other hand, imitation learning (IL) learns effectively in sparse-rewarded tasks by leveraging the existing expert demonstrations. In practice, collecting a sufficient amount of expert demonstrations can be prohibitively expensive, and the quality of demonstrations typically limits the performance of the learning policy. In this work, we propose \textit{Self-Adaptive Imitation Learning (SAIL)} that can achieve (near) optimal performance given only a limited number of sub-optimal demonstrations for highly challenging sparse reward tasks. \SAIL bridges the advantages of IL and RL to reduce the sample complexity substantially, by effectively exploiting sup-optimal demonstrations and efficiently exploring the environment to surpass the demonstrated performance. Extensive empirical results show that not only does \SAIL significantly improve the sample-efficiency but also leads to much better final performance across different continuous control tasks, comparing to the state-of-the-art.  
   % \lipsum[1]
\end{abstract}

%!TEX root = ../../main.tex
\section{Introduction}
Recent years witnessed the tremendous success of Reinforcement Learning (RL) in various tasks such as game playing \cite{mnih2015human,silver2017mastering} and robotics control \cite{gu2017deep}. Notably, RL has been advantageous in learning sequential decision-making problems with simulated environment, where massive samples with dense feedbacks can be accessed at a negligible cost. However, it is challenging to upscale RL techniques to practical applications, due to its dependence on dense reward signals to learn a long term goal. In the applications where reward feedbacks are highly sparse, RL agents may suffer from high sample-complexity, as it will struggle to connect a long sequence of actions to a delayed reward received in the far future.
Especially, for tasks with high-dimensional state-action spaces and long horizons, the learning policy may spend extremely long time exploring randomly before reaching any state with meaningful rewards.  

To learn tasks with highly sparse rewards, one promising direction is to exploit prior knowledge of expertise to facilitate learning. For example, recent advances of Imitation Learning (IL) can effectively provide remedies even when then environment rewards are unavailable, by referencing the expert demonstrations~\cite{ziebart2008maximum,ho2016generative,kostrikov2018discriminator,kostrikov2019imitation,vevcerik2017leveraging} or policies~\cite{ross2011reduction,sun2017deeply}.  
%For example, recent advances of Imitation Learning (IL) and {Learning from Demonstrations (LfD)} can effectively reduce the sampling size for efficient exploration, by referencing the expert demonstrations~\cite{ho2016generative,hester2018deep,kostrikov2018discriminator} or (near) oracle policies~\cite{ross2011reduction,sun2017deeply}. 
%The state-of-the-art IL approaches recover the expert policy by encouraging matching of state-action distributions between the expert policy and the learned policy~\cite{ziebart2008maximum,ho2016generative,kostrikov2019imitation}. 
%
Despite their success, a major limitation of such IL approaches is that the asymptotic performance of their learned policies are bounded by the given expert. As a result, when the provided demonstrations are sub-optimal, which is a more practical and challenging scenario, the IL approaches will render us a sub-optimal policy. 
%Other approaches of leveraging expertise, such as \textit{Learning from Demonstration (LfD)} \cite{kim2013learning,hester2018deep,nair2018overcoming}, although are not necessarily bounded by the demonstrations, still depends on timely reward feedbacks to learn a meaningful policy.  
%
%Another approach that leverages the behavior of experts for efficient exploration is {Learning from Demonstration (LfD)}. Combined with the value-based RL framework, LfD works by putting demonstration data in a replay buffer and treated them as self-generated data to improve sample efficiency~\cite{vevcerik2017leveraging,hester2018deep}. LfD generally needs a large number of expert demonstrations to conduct efficient learning. Moreover, most LfD approaches still require dense reward feedbacks from the environment for effective learning.

In this paper, we formally consider the problem setting where the reward feedbacks for a task is \textit{highly sparse}, and the RL agent only has access to a limited number of \textit{sub-optimal} demonstrations. In this scenario, existing IL and RL approaches will struggle to reach optimal performance, as they either depend on high-quality demonstrations or timely reward signals. 
%Our goal is to combine the merits of RL and IL to address this problem setting, by exploiting the prior knowledge from sub-optimal demonstrations that are easier to access in practice, while preserving the chance to explore for better policies guided by sparse feedbacks from the environment.
Our goal is to combine the merits of RL and IL to address this problem setting, by exploiting the sub-optimal demonstrations that are easier to access in practice, while preserving the chance to explore for better policies guided by the sparse environment feedbacks.

Towards this goal, we propose \textit{Self-Adaptive Imitation Learning (SAIL)}, an off-policy imitation learning approach that systematically adapts the learning objective to strike a balance between exploitation and exploration.
More concretely, we formulate the learning objective as exploration-driven IL. On one hand, it encourages distribution matching between the teacher policy and the current learning policy; on the other hand, it encourages the current policy to deviate from its previously learned predecessors for better exploration. 
Furthermore, to surpass the bounded performance of IL, we design an adaptive approach that carefully screens the superior self-generated trajectories based on highly sparse rewards and uses them to replace the sub-optimal demonstrations. Such treatment effectively constructs a dynamic target distribution that gradually leads to optimal performance.  
We conduct extensive empirical study and show that the proposed \SAIL achieves significant improvement in terms of both sample-efficiency and final performance across a set of benchmark tasks. 
 
% !TEX ROOT=../../main.tex
\section{Background}
\subsection{Markov Decision Process} \label{sec:preliminaries}
We consider a standard Markov Decision Process defined as a tuple $ \cM=(\cS, \cA, \cT, r, \gamma, \mu_0, \cS_0)$, where $\cS$ and $\cA$ are the state and action space;
$\cT(s'|s,a)$ denotes the transition probability from state $s$ to $s'$ upon taking action $a$; 
$r(s, a)$ is the reward function;
$\gamma$ is a discounted factor;
$\mu_0$ is the initial state distribution with $s_0 \sim \mu_0$.
$\cS_0$ is the set of terminate states or \textit{absorbing states}. 
Any absorbing state always transits to itself and yield zero rewards \cite{sutton2018reinforcement}. 
We define the return for a trajectory $\tau=\{(s_t,a_t)\}_{t=0}^\infty$ as $R_t = \sum_{k=t}^{\infty} \gamma ^{k-t}r(s_k,a_k)$.
For an episodic task with finite horizon, its return can be written as $R_t = \sum_{k=t}^{T} \gamma ^{k-t}r(s_k,a_k)$, where $T$ is the number of steps to reach the absorbing state.
The purpose of reinforcement learning (RL) is to learn a policy $\pi:~ \cS \to \cA $ that maximizes the expected return.

An important notation throughout this paper is the \textit{occupancy measure} of a policy $\pi$ \cite{ho2016generative}, defined as: 

\begin{center}
$\mu^\pi(s,a)=\sum_{t=0}^{\infty}\gamma^t Pr \big(s_t=s,a_t=a| s_0\sim \mu_0, a_t \sim \pi(s_t), s_{t+1} \sim \cT(s_t,a_t)\big)$.
\end{center}

We also define $\dP$ as the \textit{normalized stationary state-action distribution} of $\pi$, with $\dP(s,a)= (1 - \gamma) \mu^\pi(s,a)$.
Without ambiguity, we refer to \textit{density} and \textit{normalized stationary state-action distribution} interchangeably in the following paper.
From a different perspective, the objective of RL can be equivalently formulated as learning a policy $\pi$ that maximizes the expectation of rewards over $\dP$:
\begin{align} \label{obj:rl}
   \max\nolimits_\pi ~\eta(\pi) := \cE_{(s,a) \sim \dP}\big[r(s,a)\big]. 
\end{align}
%%%%%%%%%%%%%%%%%%%%%%%%%%%%%%%%%%%%%%%%%%%%%%%%%%%%%%%%%%%%%%%%%%%%%%%%%%%%%
\subsection{Adversarial Imitation Learning}
{Generative Adversarial Imitation Learning (GAIL)} addresses the imitation learning (IL) problem from the perspective of distribution matching \cite{ho2016generative}. Given a set of (near) optimal demonstrations $\cR_E$ from an unknown expert policy $\pi_E$, 
GAIL aims to learn a policy $\pi$ that minimizes the Jensen-Shannon divergence between $\dP$ and $d_E$:
\begin{align*}
   \arg\min\nolimits_{\pi} D_{JS}[\dP, \dE] - \lambda H(\pi),
\end{align*}
where $\dP$ and $d_E$ are the stationary state-action distributions for policy $\pi$ and $\pi_E$, and $H(\pi)$ is an entropy regularization inspired by maximum entropy Inverse Reinforcement Learning (IRL) \cite{ziebart2008maximum,ziebart2010modeling}. 

GAIL obtains this objective by leveraging a saddle-point optimization scheme: it jointly trains a discriminator $D$ that approaches to $D_{JS}[d_
\pi, d_E]$, and a policy $\pi$ that reduces the divergence. This learning process can be formulated by the following minimax objective:
\begin{align*} %\label{obj:gail}
   \min\nolimits_\pi \max\nolimits_{D} \cE_{\dP}[\log(1 - D(s,a))] + \cE_{d_E}[\log( D(s,a))].
\end{align*} 
Specifically, for a fixed policy $\pi$, $D$ is trained to distinguish state-actions sampled from $\pi$ and the expert policy, and outputs of an optimized discriminator $D^*$ satisfies \cite{goodfellow2014generative}:
% JZ: the teacher has not been defined. I assume you mean the demonstrations?  
$ D^*(s,a) = \frac{d_E}{d_E + \dP} $.
$\pi$ is trained in an on-policy fashion with a shaped reward function: $r'(s,a) = -\log(1-D(s,a))$.
\cite{ho2016generative} shows that optimizing the accumulated rewards of $r'(s,a)$ performs density matching between $\dP$ and $d_E$. 
%$d_E$ is approximated using samples from $\cR_E$. 
In practice, sampling from $\dP$ is obtained by \textit{on-policy} interactions with the environment. %This is also the reason that most extended work of GAIL applies to on-policy algorithms \cite{kang2018policy,fu2017learning}. 

\section{Problem Setting}
\label{sec:prob}
In this paper, we address the problem of learning in a \textit{sparse} environment with only \textit{episodic} feedbacks: dense reward signals are unavailable in this environment, except for an episodic reward provided when the task is terminated. We denote samples from such environment as $\{s,a,s',i,r_e\}$; $i$ is an indicator of whether a terminal(absorbing) state is reached:
i.e., $i(s,a,s') = 1$ if $s' \in \cS_0$ and 0 otherwise;
% \vskip -0.5cm
% \begin{equation}
%    i(s,a,s') =
%      \begin{cases}
%        1 & \text{if $s' \in \cS_0$},\\
%        0 & \text{otherwise}. \nonumber
%      \end{cases}       
% \end{equation}
% \vskip -0.2cm
%
$r_e$ denotes the episodic reward, where: 
%$r_e(s_t,a_t, s_{t+1}) = \sum_{i=0}^{t} r(s_t,a_t)$
%if $s_{t+1} \in \cS_0$ and 0 otherwise. 
 \begin{equation}
    r_e(s_t,a_t, s_{t+1}) =
      \begin{cases}
        \sum_{i=0}^{t} r(s_i,a_i) &\text{if $s_{t+1} \in \cS_0$}, \\
        0 & \text{otherwise.} \nonumber
      \end{cases}       
  \end{equation}

To alleviate the learning difficulty, the policy is allowed to leverage the prior knowledge of a limited number of demonstrations from a \textit{sub-optimal} teacher. 
This problem setting covers an extended variety of real-world scenarios, and intuitively the availability of sub-optimal demonstrations is very helpful in assisting the learning of highly sparse-rewarded tasks, as done by human learning.

To sum-up, we start learning with a random policy $\pi_0$, a teacher replay buffer with sub-optimal demonstrations $\RE$ sampled from an unknown teacher policy $\pE$, and a self replay buffer $\RB$ to store transitions generated by the policy during learning. $\RB$ is initialized with random transitions: $\RB=\{(s,a,s',i,r_e)\sim \pi_0\}$, with $i$ and $r_e$ defined above. 

To ensure that the suboptimal teacher can provide informative guidance, 
it needs to satisfy a reasonable requirement that $\pE$ performs at least better than a random policy: %Formally, we make the following practical assumption:
% {\assumption{  
\begin{assumption}[Quality of Teacher Demonstrations] 
$$ \exists\ \delta > 0, \st\ \cE_{(s,a) \sim \dE}[r(s,a)] - \cE_{(s,a)\sim d_{B}}[r(s,a)] > \delta.$$
\vskip -0.4 cm
\label{assumption:expert}
\end{assumption}

In this paper, we slightly abuse $\dE$ to denote an estimate of the stationary state-action distribution derived from $\pE$, which is approximated from the demonstration data $\RE$.
%, an approach commonly adopted by prior arts \cite{ziebart2008maximum,fu2017learning,ho2016generative}.
%
We impose Assumption~\ref{assumption:expert} only for initial learning stages, as our goal is to learn a policy $\pi$ which surpasses the teacher to reach near-optimal performance. 
%efficiently recover the teacher's performance based on its demonstrations, and gradually surpass the teacher to reach near-optimal performance via off-policy reinforcement learning. 
 
This problem setting, though widely applicable to the real world, brings significant challenges to existing RL and IL approaches.
Effective with either dense rewards or expert demonstrations, prior arts could struggle to learn expert policies when both components are unavailable.
On one hand, model-free RL depends on dense environment rewards for efficient exploration. 
%When the environment feedback is highly sparse for long horizon tasks, it may cause enormous redundant exploration;
% before collecting any reasonable trajectories;
%
On the other hand, IL approaches heavily rely on high-quality demonstrations to recover expert performance. 
%Conducting IL on imperfect demonstrations leads to suboptimal performance of the learning policy.

To tackle this challenging problem setting, we need a novel approach which efficiently boosts learning with sub-optimal teacher demonstrations, and at the same time enables efficient exploration, by leveraging highly sparse rewards from the environment.  
% !TEX ROOT=../../main.tex
\section{Methodology} 
\subsection{Exploration-Driven Imitation Learning}  
 
\begin{figure}[t!] 
\centering
	\includegraphics[width=0.4\textwidth]{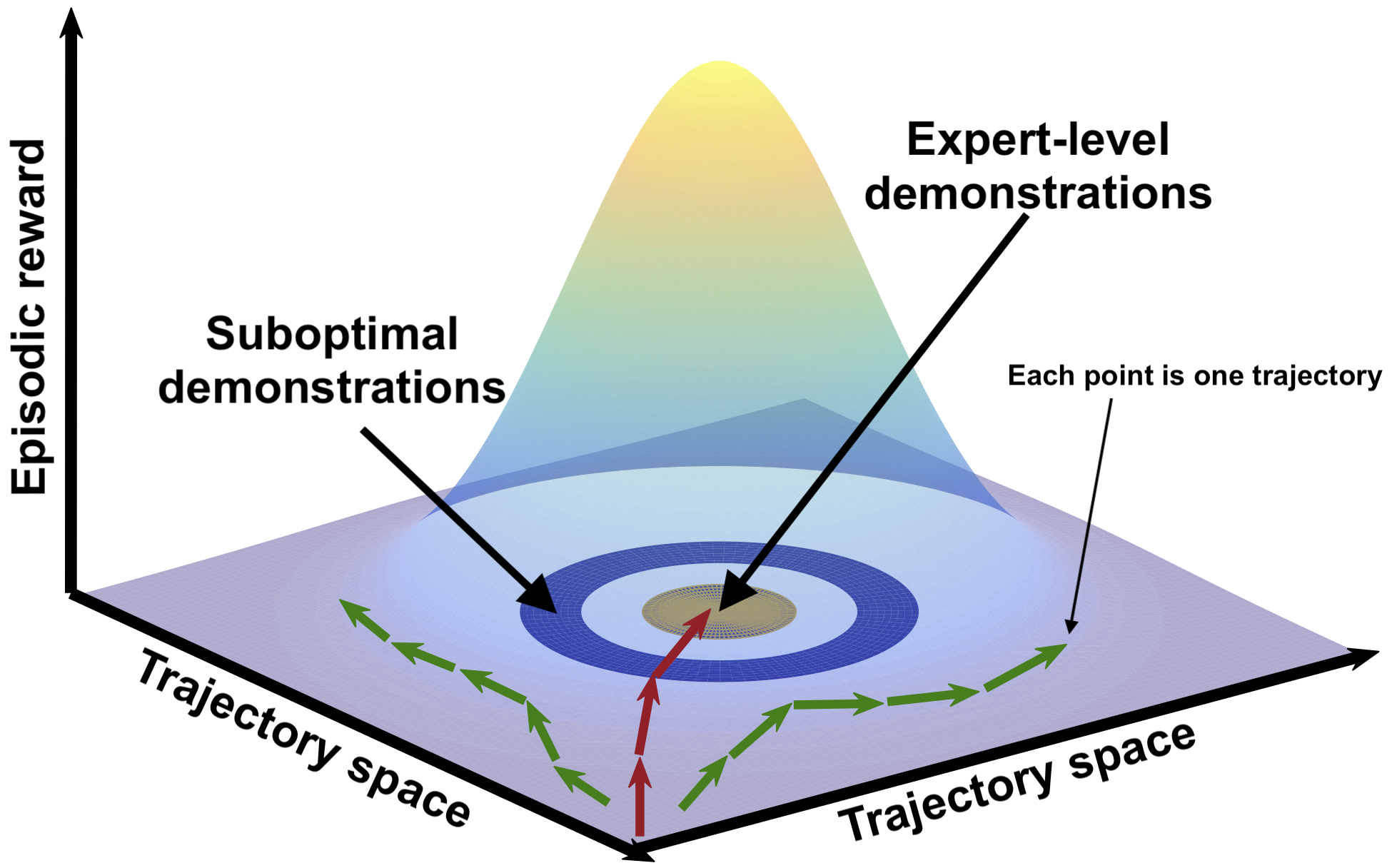} 
	\begin{center}
	%\caption{The proposed SAIL can quickly reach the high-reward region in the trajectory space by utilizing the suboptimal demonstrations. The
	%red arrows denote the navigation of SAIL in the trajectory space, where
	%each point in the plane represents one trajectory. 
	%Therefore, it explores more efficiently than the random exploration (green arrows) in sparse-rewarded tasks.}
	\caption{Each point in the plane represents one trajectory. The red arrows denote the navigation process guided by our exploration-driven IL objective, which approaches to teacher's density distribution and deviates from previous learned densities. It explores more efficiently to reach expertise, compared with random explorations in sparse-rewarded tasks (green arrows).}
	\label{fig:sail-ill}  		
	\end{center}
	\vskip -0.3in
\end{figure}  
For the purpose of deriving a solution benefiting from both  \textit{highly-sparse} rewards and \textit{sub-optimal} demonstrations, we propose a novel objective which strikes a balance between exploration and imitation learning, formulated as follows:
%which encourages efficient exploration in the case of sparse rewards, while fully utilizes the imperfect demonstrations provided (as illustrated in Figure~\ref{fig:sail-ill}).
% 
\begin{align} 
  \hspace{-1mm}\max_\pi ~ J(\pi) := - D_{\KL}[d_\pi || \dE] + D_{\KL}[d_\pi || d_{B}],   \label{eq:obj-primal1}
   % =  &~\cE_{(s,a) \sim d_\pi}[-\log \frac{d_\pi(s,a)}{\dE(s,a)}]  +  \log \frac{d_\pi(s,a)}{\dB(s,a)}] \nonumber\\
   % & = \max_{\pi} \cE_{(s,a) \sim d_\pi}[\log \frac{\dE(s,a)}{\dB(s,a)}]. \label{eq:obj-primal2}
\end{align}
 % }}
where $D_{KL}$ denotes the KL-divergence between two distributions, with $D_{KL}[P||Q] = \cE_{x\sim p(x) } \log \frac{p(x)}{q(x)}$.

The objective in \eq{\ref{eq:obj-primal1}} can be interpreted as joint motivations for imitation and exploration.
The first term $- D_{\KL}[d_\pi || \dE]$ encourages density distribution match between $d_\pi$ and $\dE$. The second term $D_{\KL}[d_\pi || d_{B}]$, though could be counter-intuitive at first sight, serves as an objective for self-exploration. 
Since $\dB$ is the normalized state-action distribution derived from a mixture of previously-learned policies, maximizing $D_{\KL}[d_\pi || d_{B}]$, \ie $\cE_{(s,a)\sim d_\pi}[\log\frac{d_\pi(s,a)}{\dB(s,a)}]$, is in favor of visiting state-actions that are rarely seen by previously learned policies.
Combined with the imitation term, this self-exploration is guided in the direction that decreases $D_\KL[d_\pi || \dE]$.

% \subsection{Benefits of exploration-driven objective given sub-optimal demonstrations:}

We can rewrite the objective in \eq{\ref{eq:obj-primal1}} as the following equivalent problem: 
\vskip -0.1in
\begin{align} 
  \max_\pi ~ J(\pi) := \cE_{(s,a) \sim d_\pi}\left[\log \frac{\dE(s,a)}{\dB(s,a)}\right], \label{eq:obj-primal2}
\end{align} 
\vskip -0.1in

which provides two important insights into the proposed learning objective. 
%
% To get more insight on the proposed learning objective, 
% we equivalently reformulate the objective as follows:
% \vskip -0.5 cm
% \begin{align} 
%   \max_\pi ~ J(\pi) := \cE_{(s,a) \sim d_\pi}\left[\log \frac{\dE(s,a)}{\dB(s,a)}\right]. \label{eq:obj-primal2}
% \end{align}
% \vskip -0.3 cm
%
%Advantages of optimizing the above objective are two-folds. 
First, it provides guidance for policy $\pi$ to selectively build its \textit{support}, leading to the acceleration of the IL process. 
One can consider maximizing $\cE_{(s,a) \sim d_\pi}\left[\log \tfrac{\dE(s,a)}{\dB(s,a)}\right]$ as a process of policy selection: for state-actions where the teacher has visited more frequently than the previously-learned policies, $\pi$ is encouraged to build positive densities on those state-actions, leading to $d_\pi(s,a) > 0$ where $\dE(s,a) > \dB(s,a)$. 
Intuitively, this process implies that we trust the teacher more than the previously learned policies.
Given Assumption~\ref{assumption:expert}, this is a fair policy search scheme in early learning stages.
After policy $\pi$ reaches the teacher's performance, we derive an paradigm to adaptively adjust the teacher's demonstration buffer, which will effectively improve the upper-bound of the teacher's performance, and naturally keeps Assumption \ref{assumption:expert} tenable. We will elaborate this process later in this paper. An illustration of optimizing objective in \eq{\ref{eq:obj-primal2}} is shown in Figure \ref{fig:sail-ill}.

On the other hand, our proposed objective encourages exploration, as opposed to an IL objective that solely encourages distribution match between $d_\pi$ and $\dE$:
\begin{align*}  
  \max\nolimits_\pi J_{\text{IL}}(\pi) := - \text{Dist}[d_\pi || \dE], %\tilde{J}_{\dE}(\pi) =
%    =  \max_\pi &\cE_{(s,a) \sim d_\pi}[\log \frac{\dE(s,a)}{d_\pi(s,a)}] + \lambda H(\pi) \nonumber
\end{align*} 
% $\max\nolimits_\pi - D[d_\pi || \dE]$,
where $\text{Dist}$ is an arbitrary divergence measure, such as the Jensen-Shannon divergence or KL-divergence \cite{nowozin2016f}. 
%Popular divergence measures can be KL-divergence or JS-divergence \cite{nowozin2016f}.
%
An optimal solution to the IL objective is a policy that exactly recovers the teacher's density distribution with $d_\pi^*= \dE$ \cite{ziebart2008maximum}, which leads to an imperfect policy when the teacher's performance is far from optimal.
Moreover, optimizing $J_{\text{IL}}(\pi)$ alone restricts $\pi$ from further exploring density distributions that deviate from $\dE$.
We verify by empirical studies in Section \ref{sec:abstudy} that, the objective of $\max_\pi E_{d_\pi}[\log(\frac{\dE}{\dB})]$ achieves more efficient exploration, as compared to $\max_\pi E_{d_\pi}[\log(\frac{\dE}{d_\pi})]$, a pure imitation-driven objective.

%\judycom{(add Remark: objective is same as dynamic reward shaping)}

While the objective in \eq{\ref{eq:obj-primal1}} is appealing in terms of combining the merits of both IL and exploration, directly optimizing it presents significant challenges. First, the density ratio of $\log\frac{\dE}{\dP}$ is hard to estimate, especially when both replay buffers $\RB$ and $\RE$ are dynamically updated. 
Second, it requires the state-action to be sampled from 
the current policy to get expectations over $\dP$, which will encounter the same sample-inefficiency issue faced by on-policy RL. 

Next, we propose practical alternative objectives to systematically resolve the aforementioned issues.

% !TEX ROOT=../../main.tex
\subsection{Off-Policy Adversarial TD Learning} \label{sec:lfd-theory}

\begin{remark}[]
   The proposed objective in \eq{\ref{eq:obj-primal2}} is equivalent to a max-return RL objective (as defined in \eq{\ref{obj:rl}}), with $\log\frac{\dE}{\dB}$ in place of the environment rewards. 
\end{remark} 

Based on the above insight, we can connect the primal objective to {Temporal-Difference (TD) learning}, and solve it by leveraging an \textit{actor-critic} method.
Especially, we learn a $Q$-function (critic) to estimate an expected return by minimizing its TD errors, and a policy $\pi$ (actor) is learned to maximize that expectation, given $\log\frac{\dE}{\dB}(s,a)$ as rewards. 
%
%
% Instead of optimizing \eq{\ref{eq:obj-primal2}} directly, the policy (actor) is trained to maximize the expected Q values.

To obtain $\log\frac{\dE}{\dB}$, we built upon the prior arts~\cite{ho2016generative} to learn a discriminator $D$ that maximizes the following objective:
% JZ: The notation D is used to refer both distance and discriminator, which are not good. consider changing the distance one using $\text{Dist}$ ($d$ is used for distribution). 
{\small{
\begin{align} \label{eq:discriminator}
\max_{D:\cS \times \cA \to (0,1) } J_D := \cE_{\dB}[\log(1 - D(s,a))] + \cE_{\dE}[\log( D(s,a))],
\end{align}
}}
%
%\vskip -0.5 cm
$D$ is trained to distinguish the self-generated data from $\RB$ and the teacher demonstrations from $\RE$.
When the discriminator is trained to reach optimality, its output satisfies
$D^*(s,a)=\frac{\dE(s,a)}{\dE(s,a) + \dB(s,a)}$ \cite{goodfellow2014generative}. The output of $D$ with a constant shift, which we found is empirically effective, is used to approximate a reward function:
%{\small{
\begin{align*}
r'(s,a) = -\log(1-D(s,a)) \approx \log(\tfrac{\dE}{\dB}+1).
\end{align*}
%}}
%\vskip -0.1 in
% $$r'(s,a) = -\log(1-D(s,a)) \approx \log(\frac{\dE}{\dB}+1).$$
% \vskip -0.2 cm 
In the initial training stage, a well-trained discriminator renders higher rewards to teacher demonstrations with $D(s,a) \to 1$, and lower rewards for self-generated samples with $D(s,a) \to 0$. 

Based on the shaped reward function, we learn a deterministic policy which confuses the discriminator to maximize the shaped returns. To get around the issue of sample inefficiency, we adopt an \textit{off-policy} learning framework. Our objective is accordingly altered to maximize the expectation of $Q$-value over state distributions of a behavior policy $\beta$ \cite{silver2014deterministic}:
\begin{align*}
   J_\beta(\pi_\theta) =& \int _s d_\beta(s) Q(s,\pi_\theta(s)) d_s, 
   = \cE_{s \sim d_\beta(s)}[Q(s, \pi_\theta(s))],
\end{align*}
where $d_\beta(s)$ is the normalized stationary \textit{state} distribution of $\beta$, which is analogous to the \textit{state-action} distribution defined in Section \ref{sec:preliminaries},
and the $Q$-function is built based on shaped rewards:
\begin{align*}
   Q(s,a)=r'(s,a) + \gamma \cE_{s'\sim \cT(s'|s,a),a'\sim \pi(s')}[Q(s',a')],
\end{align*}  
\vskip -0.1 in
where we use deep neural networks to approximate both $Q$ and $\pi$, parameterized by $\phi$ and $\theta$, respectively.
$Q$ is learned to minimize the TD error, averaged by an off-policy stationary distribution generated by the behavior policy $\beta$:
\begin{align} \label{eq:Q-obj}
   \min_\phi J_\beta{Q_\phi} := \cE_{s\sim d_\beta,a\sim \beta}[(Q_\phi(s,a)-y)^2]. 
\end{align} 
\vskip -0.1 in
We adopt a target network $\bar{Q}$ to stabilize training. Accordingly, $y = r'(s,a) + \gamma \bar{Q}(s',\pi_\theta(s'))$.

The policy-gradient for the actor can be derived as \cite{lillicrap2015continuous}:
\begin{align} \label{eq:pi-obj}
   \nabla_\theta J_\beta(\pi_\theta) \approx \cE_{s\sim d_\beta}[\nabla_\theta \pi_\theta(s) \nabla_a Q_\phi(s,a)|_{a=\pi_\theta(s)}].
\end{align} 
\vskip -0.1in
For regular off-policy RL methods, the behavior policy is instantiated as the transition dataset collected by policies learned through iterative training, a scheme which has been proved to greatly improve sample efficiency compared to \textit{on-policy} sampling.
In next Section, we introduce an algorithm which realizes our objective via abovementioned off-policy TD learning. It adopts an even more effective sampling approach that further accelerates the IL process.

% !TEX ROOT=../../main.tex
\subsection{Self-Adaptive Imitation Learning} \label{sec:algorithm}

\begin{algorithm}[tb]
    \caption{Self-Adaptive Imitation Learning}
    \label{alg:concise}
 \begin{algorithmic}
    \State Input: teacher replay buffer $\RE$, self-replay-buffer $\RB$, initial policy $\pi_\theta$, discriminator $D_w$, critic $Q_\phi$, the initial ratio of samples from teacher demonstrations: $\alpha = 0.5$. 
    %%%%%%%%%
    \For{$n=1$, $\dots$} 
    \State Sample a trajectory $\tau \sim \pi_\theta$, $\RB \leftarrow \RB \cup \tau$.
    \If{$r_e(\tau) > \min_{\tau'} \{r_e(\tau')|\tau' \in \RE\} $ }
       \State Add good trajectory to $\RE$: $\RE \leftarrow \RE \cup \tau$.
       \State Anneal $\alpha \rightarrow 0$.
    \EndIf
    % \IF{ $n \mod \code{discriminator-update}= 0$} 
    \State Update discriminator $D_w$ by optimizing \eq{\ref{eq:discriminator}}.
    % \ENDIF
    % \IF{ $n \mod \code{Q-update}= 0$}    
	\State Update critic $Q_\phi$ by optimizing \eq{\ref{eq:qgrad}}.
    % \ENDIF
    % \IF{ $n \mod \code{policy-update}= 0$}    
    \State Update policy $\pi$ via gradient accent in~\eq{\ref{eq:pigrad}}.
    % \ENDIF 
    \EndFor 
 \end{algorithmic}
 \end{algorithm}  

\begin{figure}  
	\centering 
	\includegraphics[width=0.5\textwidth]{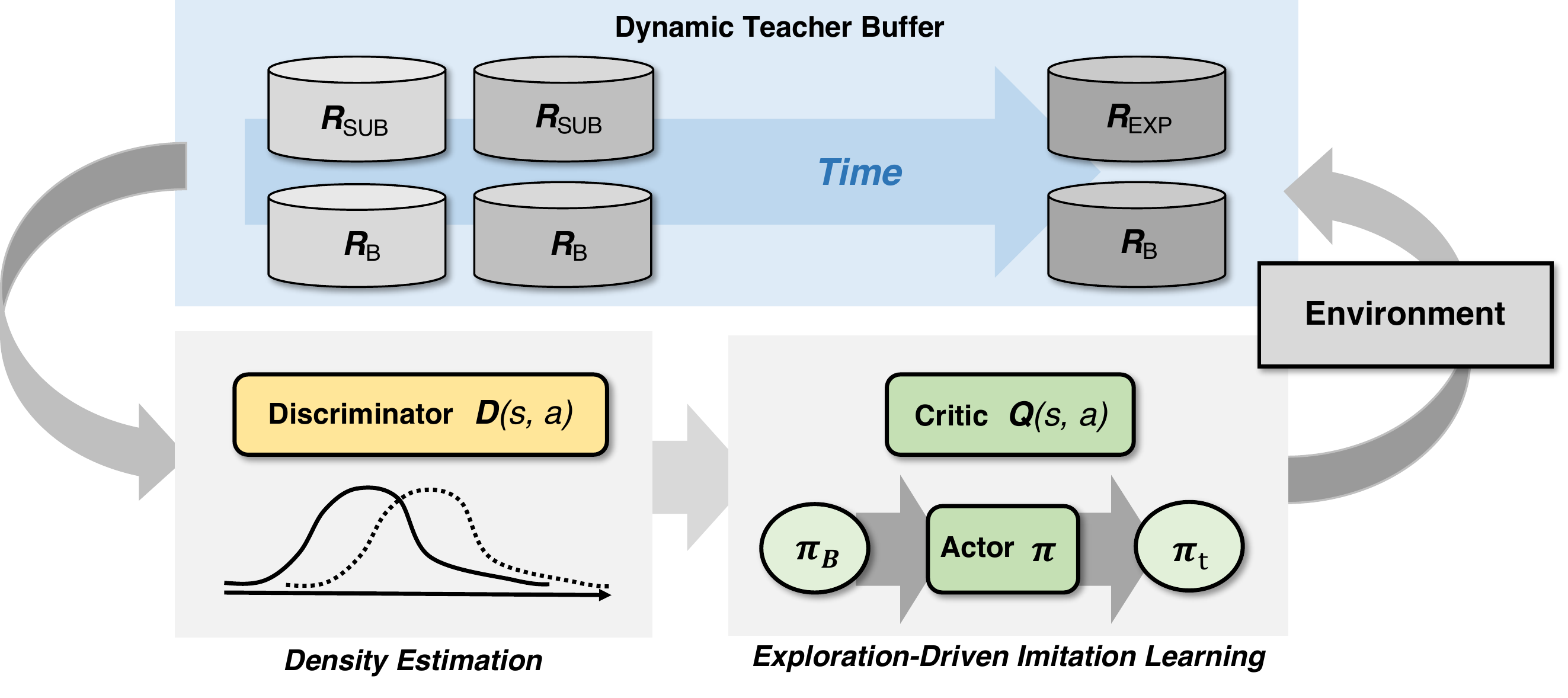}
	\caption{The illustration of the proposed Self-Adaptive Imitation Learning (\textit{SAIL}) framework.} 
    %\vspace{-0.1in}
    \label{fig:sail-framework}
    %\vskip -0.1in
\end{figure}

In this section, we introduce  \textit{Self-Adaptive Imitation Learning (SAIL)}, an algorithm based on objectives mentioned in above section.
%an algorithm that enables exploration-driven off-policy imitation learning.
%
\SAIL maintains two replay-buffers $\RE$ and $\RB$, for caching teacher demonstrations and self-generated transitions, respectively. 
It alternatively trains three components: a discriminator $D$ that serves as a reward provider, a critic $Q$ that minimizes the Bellman error based on the shaped rewards, and an actor $\pi$ that maximizes the shaped returns. 
During iterative training, \SAIL adds high-quality self-generated trajectories into the teacher demonstration buffer.
An illustration of \SAIL is shown in Figure~\ref{fig:sail-framework} and
a simplified pseudo-code is provided in Algorithm~\ref{alg:concise}. 
Due to the paper space limit, we defer the formal summarization of \SAIL to Appendix. We highlight three key aspects that make \SAIL effective:
 
(1) \textit{Leveraging sparse environment feedback to update teacher demonstration buffer $\RE$}:  
If the behavior policy collects a trajectory with episodic reward $r_e$ higher than a threshold $C_\dE$, we treat this trajectory as teacher demonstrations and store it in buffer $\RE$. The threshold $C_\dE$ is simultaneously updated as the lowest episodic score in $\RE$: 
% $$C_\dE = \min\nolimits_{r^j_e \neq 0}\{r^j_e |(s_j,a_j, \dots, r^j_e \in \RE) \}$$
$C_\dE = \min_{\tau'} \{r_e(\tau')|\tau' \in \RE\}$.
Guided by the exploration-driven objective, a student policy $\pi$ can quickly recover the teacher's performance and dynamically construct the demonstration buffer with gradually increasing quality.
  
(2) \textit{Realizing the exploration-driven objective by reshaping rewards with an off-policy discriminator}: 
Prior work such as \GAIL tackles the IL problem solely from the perspective of distribution matching.
Obtaining their objective relies on training a discriminator with \textit{on-policy} samples from the current policy $\pi$, which is sample-inefficient.
% 
%, and \textit{off-line} samples from an expert replay buffer $\cR_E$, to reshape the per-step reward as the density ratio between $d_\pi$ and $d_E$. 
%, and samples from an expert replay buffer $\RE$, respectively.
%, which reshapes the per-step reward as the density ratio between $d_\pi$ and $\dE$.
%The discriminator is fed with on-policy samples collected from the current policy $\pi$, and samples from an expert replay buffer $\RE$, respectively.
%
% Contrastively, we seek for a policy that strikes a balance between exploration and IL, by shaping the reward to guide policy $\pi$ to approach to its teacher while deviates from its predecessors.
%Especially, at $k$-th iteration, the shaped reward encourages the current policy $\pi_k$ to deviate from its predecessors 
%.$\pi_1, \pi_2, \dots, \pi_{k-1}$. 
%
To resolve this issue, we iteratively train a discriminator $D$ using \textit{off-policy} samples, whose output represents a ratio of $\frac{\dE}{\dB}$, where $\dB$ is the density distribution derived from a mixture of previously learned policies, and $\dE$ is the density distribution of the teacher.
We show via ablation studies that, compared to its on-policy counterpart, our off-policy discriminator aligns with the proposed objective and encourages more efficient exploration.
%, given only sub-optimal teacher demonstrations.

(3) \textit{Fully utilizing teacher demonstrations to boost imitation learning}:
In the same spirit of learning from demonstrations~\cite{hester2018deep,nair2018overcoming,sun2018truncated}, 
we sample from both teacher dataset in $\RE$ and self-generated dataset in $\RB$, to construct a mixed density distribution, which plays the role of $d_\beta$ in \eq{\ref{eq:Q-obj}} and \eq{\ref{eq:pi-obj}}.
%
% Perhaps one outstanding benefit of this sampling scheme is that, it
This sampling scheme naturally ensures \textit{a better policy output to remain in the support of the training dataset distribution}, which could alleviate the issue of out-of-distribution actions \cite{kumar2019stabilizing}. 
More concretely, this process can be formulated as training $Q$ and $\pi$ to optimize their objective values as follows:
% \vskip -0.2in
% \begin{align}
%     &\min_\phi J(Q_\phi) := \cE_{(s,a,s') \sim \dMIX}[(Q_\phi(s,a)-y)^2], \label{eq:qgrad}\\
%     &\triangledown_\theta J_\beta(\pi_\theta) \approx \cE_{ (s,a) \sim \dMIX }[\triangledown_\theta \pi_\theta(s) \triangledown_a Q_\phi(s,a)|_{a=\pi_\theta(s)}].\label{eq:pigrad}
% \end{align}
\begin{align}
    &\min_\phi \hat{J}(Q_\phi) := \frac{1}{N}\sum\nolimits_{i=1}^N(Q_\phi(s_i,a_i)-y_i)^2, \label{eq:qgrad}\\
    &\nabla_\theta J_\beta(\pi_\theta) = \frac{1}{N}\sum\nolimits_{i=1}^N\nabla_\theta \pi_\theta(s_i) \nabla_{a} Q_\phi(s_i,a)|_{a=\pi_\theta(s_i)}],\label{eq:pigrad}
\end{align}
 % \vspace{-1em}
%
where the state action pairs are sampled from the mixture of two buffers: $\dMIX = \alpha \dE + (1 - \alpha) \dB$. $\alpha$ is the ratio of samples from teachers demonstrations, and $y_i$ is defined same as in the objective of \eq{\ref{eq:Q-obj}}. 
In practice, we initialize $\alpha=0.5$. Once the learning policy is able to generate trajectories with performance comparable to the teacher, we gradually anneal the value of $\alpha$ to zero. 
We validate in Section \ref{sec:abstudy} that, learning from a mixture dataset from the teacher and the student is more beneficial for accelerating the initial learning stage.

%Perhaps one outstanding benefit of learning with teacher demonstrations is that it naturally ensures \textit{a better policy output to remain in the support of the training dataset distribution}. 
%

% !TEX ROOT=../../main.tex

\section{Experiments}
In this section, we study how  \SAIL performs in terms of reaching the objective of imitation learning and exploration.
We conduct extensive experiments to answer the following key questions: %1) Is \SAIL sample-efficient? 2) Can SAIL surpass the demonstration performance via exploration? 3) Which components in SAIL contribute to sample-efficiency and exploration?

    Q1. Is \SAIL sample-efficient? 
 
    Q2. Can \SAIL surpass the demonstration performance via exploration?

    Q3. Which components in \SAIL contribute to the sample-efficiency or exploration ?

%\end{itemize}

\subsection{Setup}
We implemented \SAIL on a TD3 framework \cite{fujimoto2018addressing} using open-sourced code from \textit{stable-baselines}\footnote{https://stable-baselines.readthedocs.io/en/master/}.
We tested \SAIL on 4 modified locomotion tasks simulated by OpenAI MuJoCo\footnote{https://github.com/openai/mujoco-py}: \textit{Walker2d-v2, Hopper-v2, HalfCheetah-v2}, and \textit{Swimmer-v2}.
The original tasks are in the dense-reward setting. To construct a sparse feedback environment, we omit the original rewards such that only an episodic reward is given upon the completion of a trajectory.
For each task, we generated teacher demonstrations by training a TD3 benchmark to reach sub-optimal performance, and use trajectories with triplets $(s_t, a_t, s_{t+1})$ to initialize the teacher replay buffer $\RE$. 

We compare \SAIL with 4 baselines that are mostly applicable to our problem setting: DAC~\cite{kostrikov2018discriminator}, GAIL~\cite{ho2016generative}, POfD~\cite{kang2018policy}, and Behavior Cloning (BC).
DAC is an off-policy imitation learning baseline built also on a TD3 framework, which attaches auxiliary \textit{absorbing states} to each episode to stabilize learning. 
POfD is an extension of GAIL that maximizes policy returns of combined rewards from the environment and the discriminator. The reward used by POfD can be denoted as $r'(s,a)=  r(s,a) - \lambda\log(1-D(s,a))$. Fitted into our problem setting, we use episodic rewards $r_e(s,a)$ in place of $r(s,a)$, and scale the shaped reward as $r'(s,a) = 0.1 * r_e(s,a) - \log(1-D(s,a))$. As on-policy baselines, both POfD and GAIL are built upon the TRPO framework~\cite{schulman2015trust}. All experiments are conducted using 5 random seeds, with seed numbers ranging from 1 to 5. More details about experiment implementations are provided in the supplementary document.

\subsection{Performance on Continuous Action-Space Tasks} 
We conduct comparison between \SAIL and other baselines using 1 and 4 trajectories of imperfect demonstrations, respectively. Experimental results are analyzed from the perspective of sample complexity and exploration ability. 

\begin{figure*}[ht]
    \begin{center}  
        %\vskip -0.1in
        \hskip -0.1in
        \begin{minipage}[b]{.27\textwidth}
            \centerline{\includegraphics[width=\columnwidth]{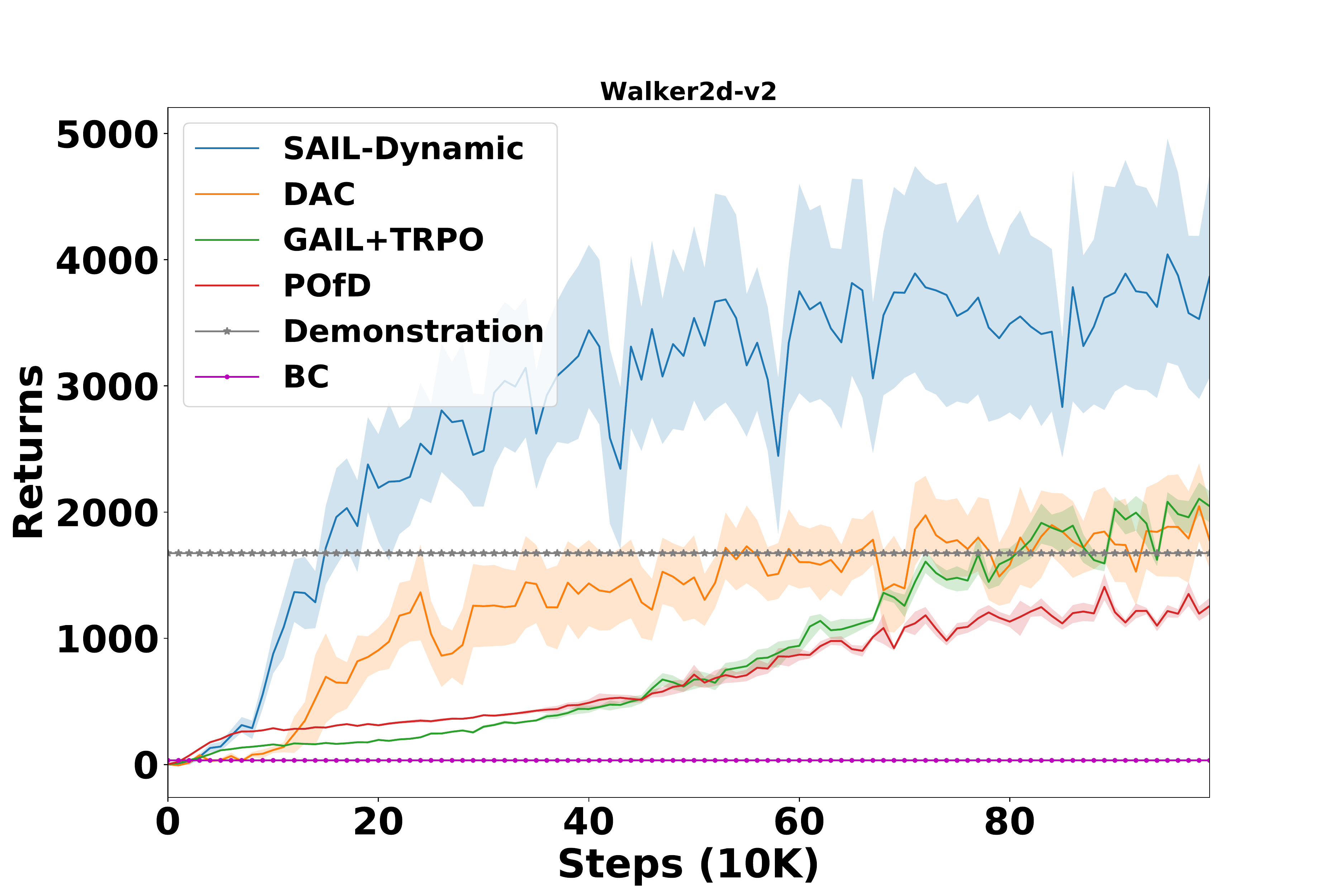}} 
            %\caption{Walker2d-v2}\label{ep4-walker}
        \end{minipage} %\hskip -0.3in %\qquad  
        \hskip -0.2in
        \begin{minipage}[b]{.27\textwidth}
            \centerline{\includegraphics[width=\columnwidth]{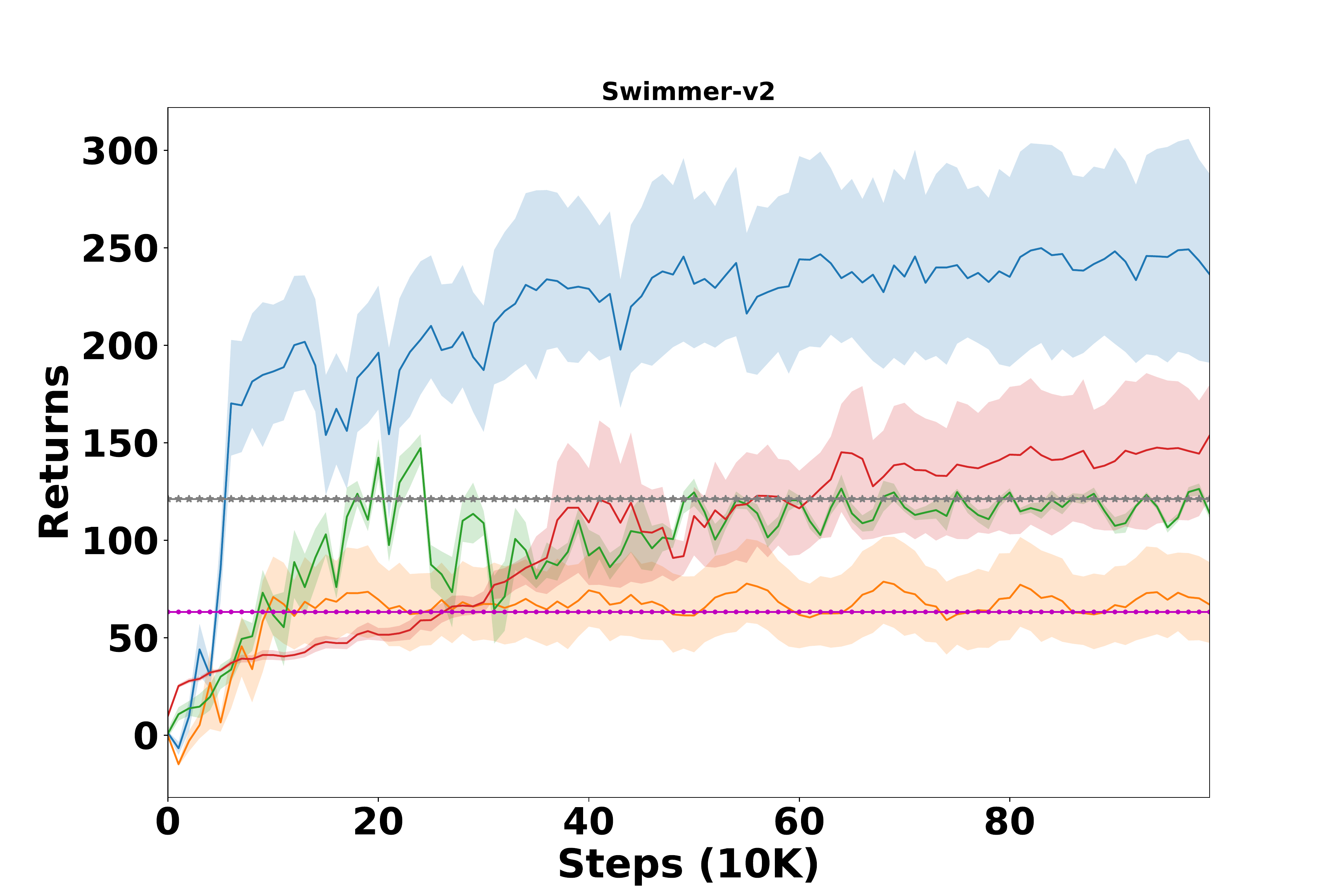}}
            %\caption{Swimmer-v2}\label{ep4-swimmer}
        \end{minipage} \hskip -0.2in
        \begin{minipage}[b]{.27\textwidth}
            \centerline{\includegraphics[width=\columnwidth]{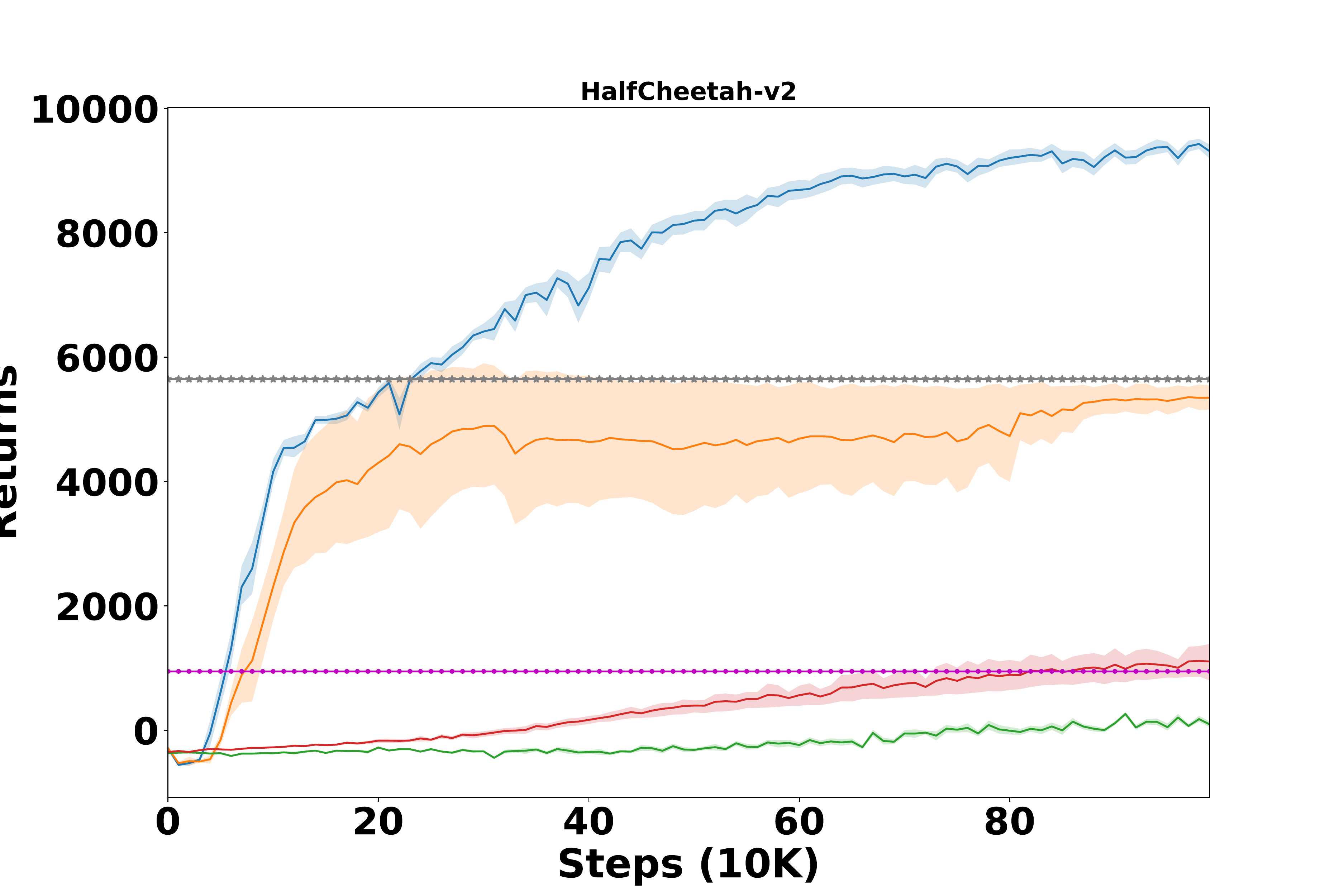}}
            %\caption{Swimmer-v2}\label{ep4-swimmer}
        \end{minipage} \hskip -0.2in
        \begin{minipage}[b]{.27\textwidth}
            \centerline{\includegraphics[width=\columnwidth]{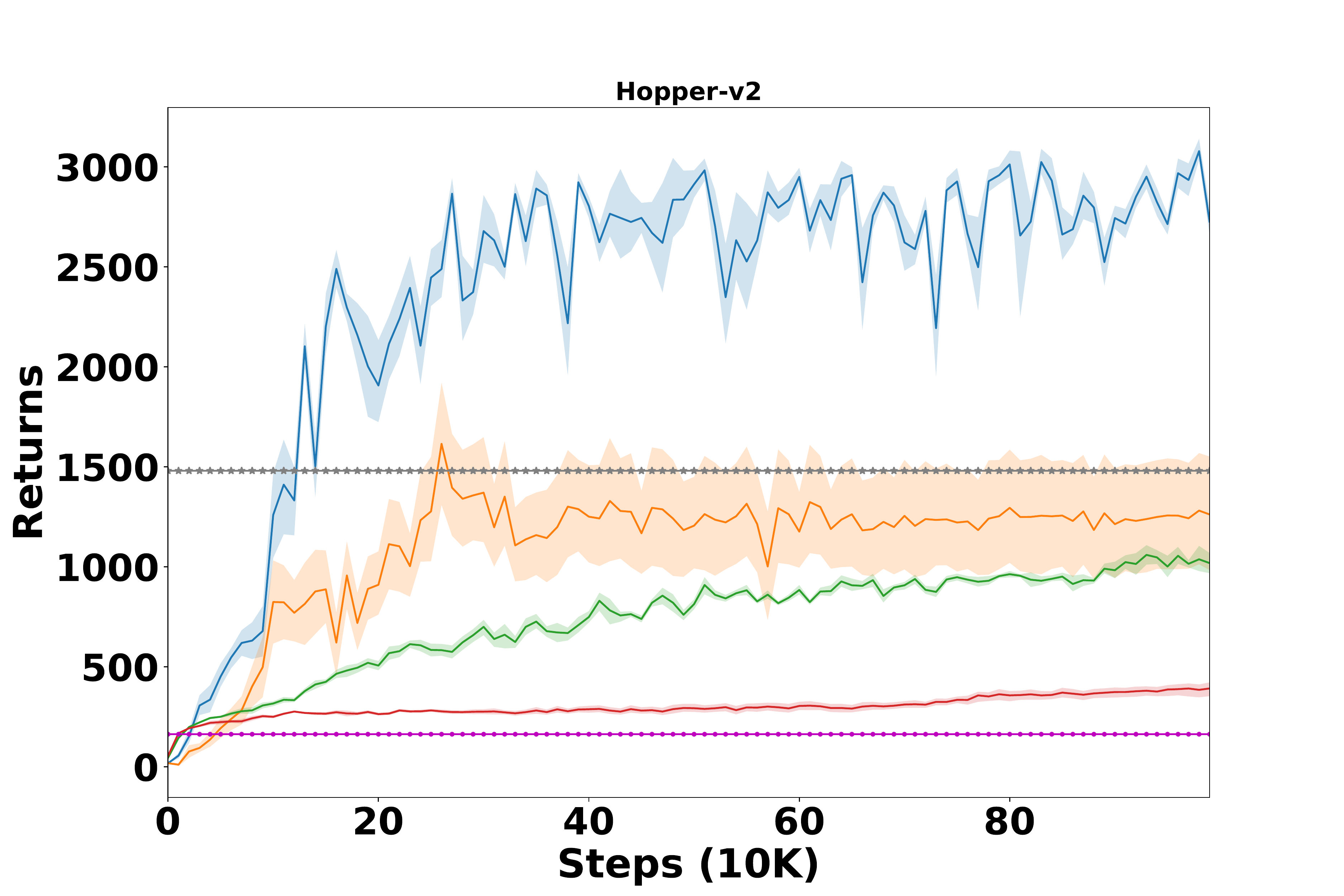}} 
        \end{minipage}%\qquad    
        %\vskip -0.1in  
    \caption{Learning curves of \SAIL and other baselines using 1 suboptimal demonstration trajectory.}
    \label{fig:ep1-sail-dynamic}
    \end{center}
    %\vskip -0.1in
\end{figure*}

\begin{figure*}[ht] 
    \begin{center}
        \hskip -0.1in %\qquad 
        \begin{minipage}[b]{.27\textwidth}
        %\begin{minipage}[b]{.33\textwidth}
            \centerline{\includegraphics[width=\columnwidth]{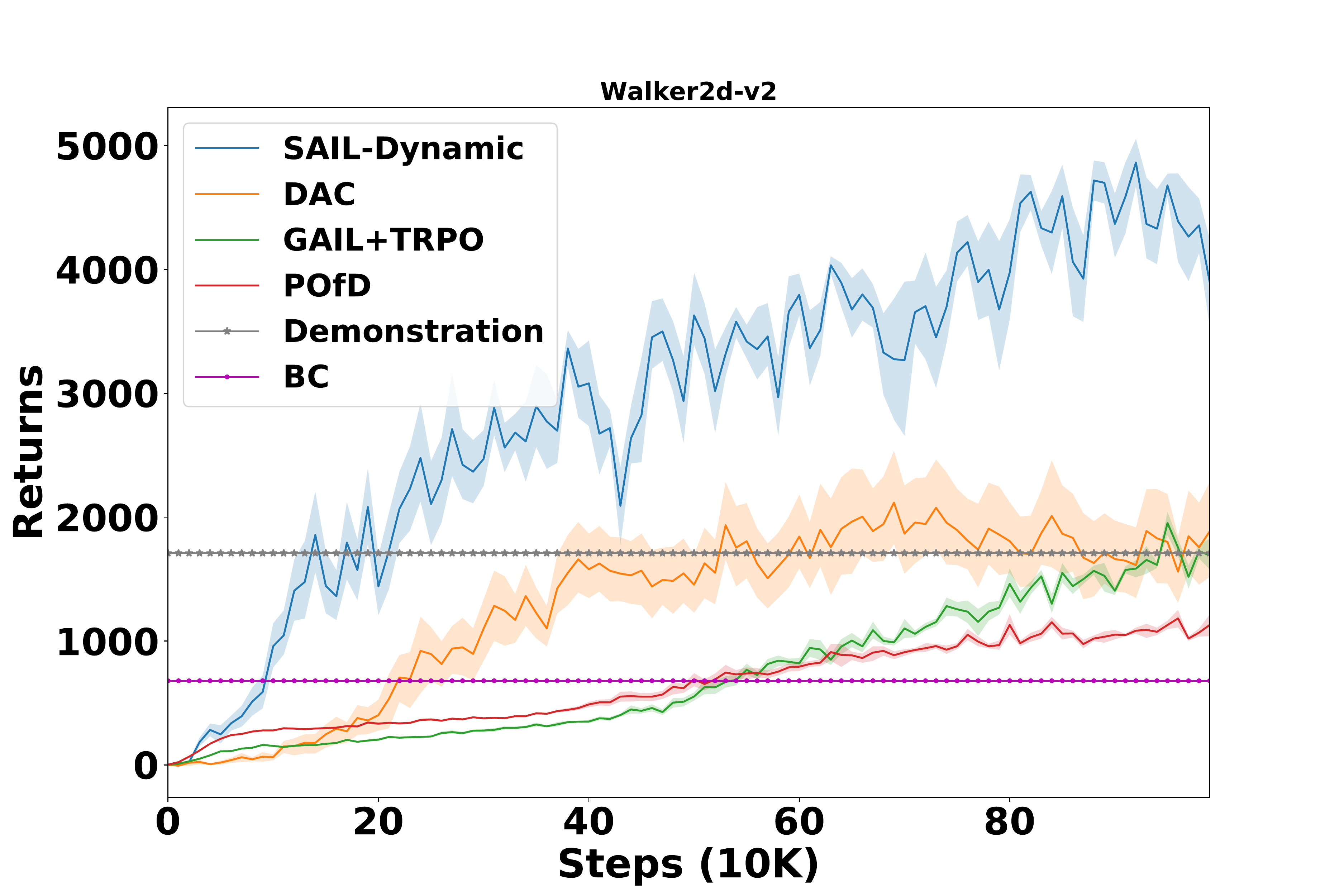}}  
        \end{minipage} 
        \hskip -0.2in %\qquad 
        \begin{minipage}[b]{.27\textwidth}
        %\begin{minipage}[b]{.33\textwidth}
            \centerline{\includegraphics[width=\columnwidth]{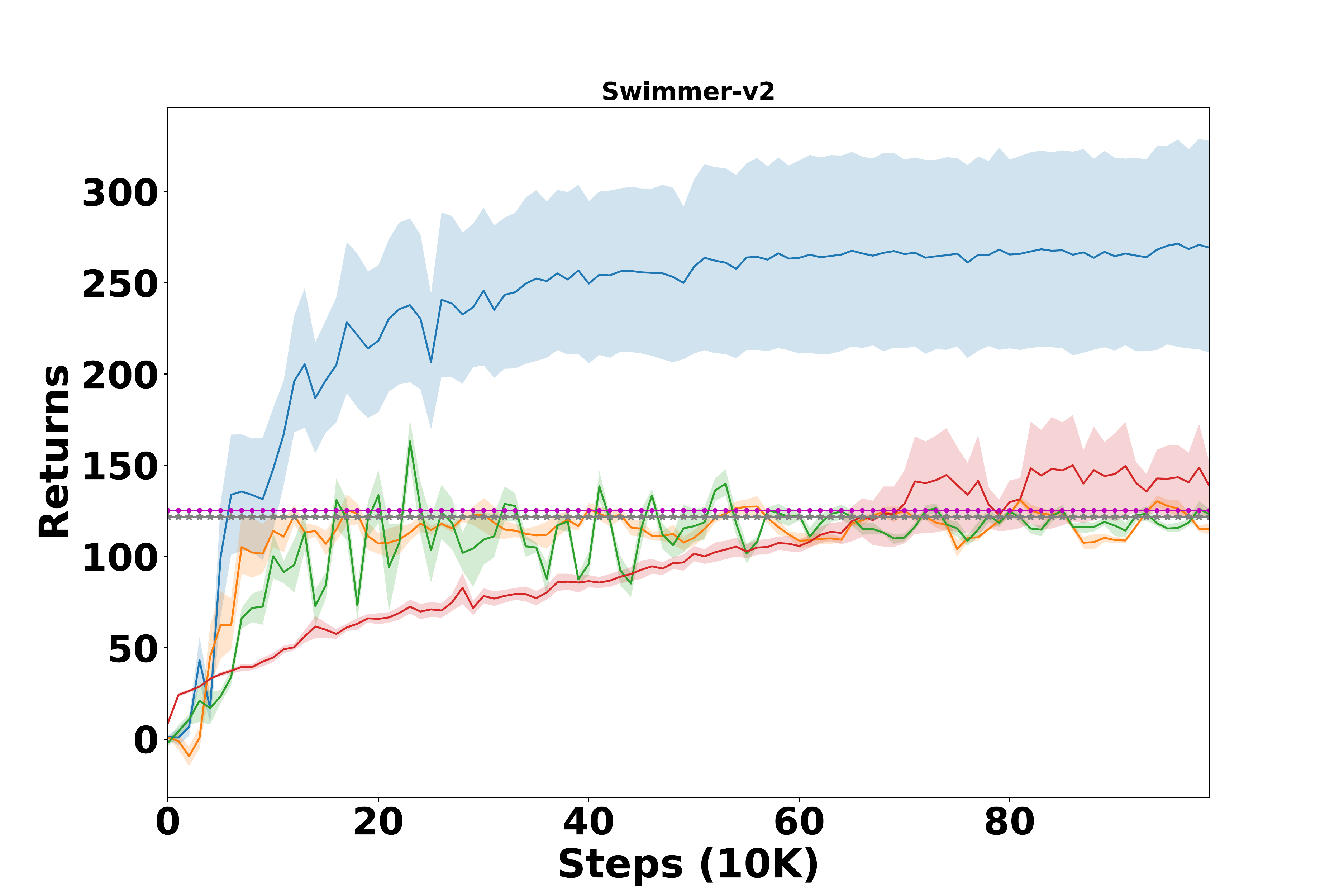}} 
        \end{minipage} %\hskip -0.3in
        \hskip -0.2in %\qquad 
        \begin{minipage}[b]{.27\textwidth}
        %\begin{minipage}[b]{.33\textwidth}
            \centerline{\includegraphics[width=\columnwidth]{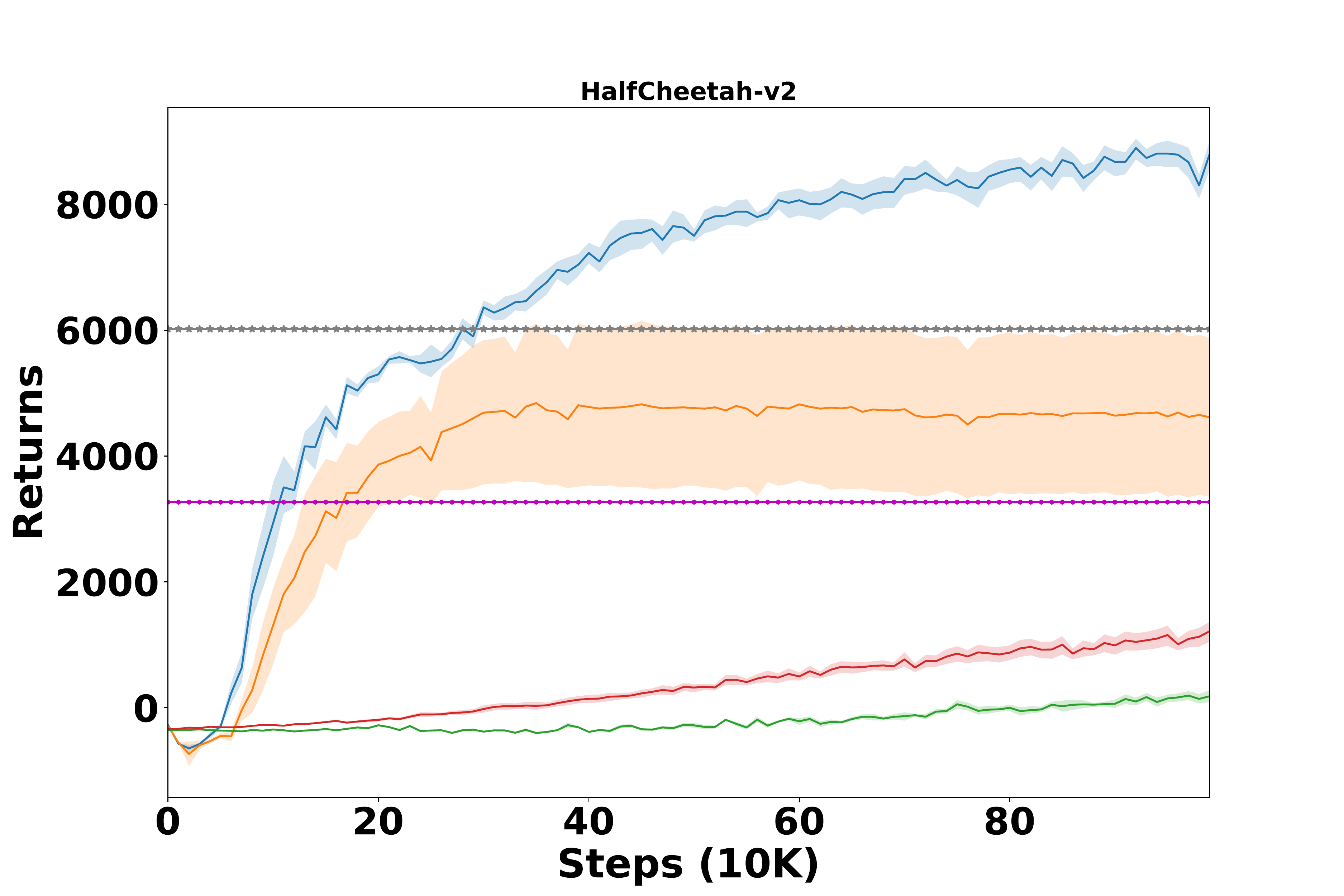}} 
        \end{minipage} %\hskip -0.2in 
        \hskip -0.2in %\qquad 
        \begin{minipage}[b]{.27\textwidth}
        %\begin{minipage}[b]{.33\textwidth}
            \centerline{\includegraphics[width=\columnwidth]{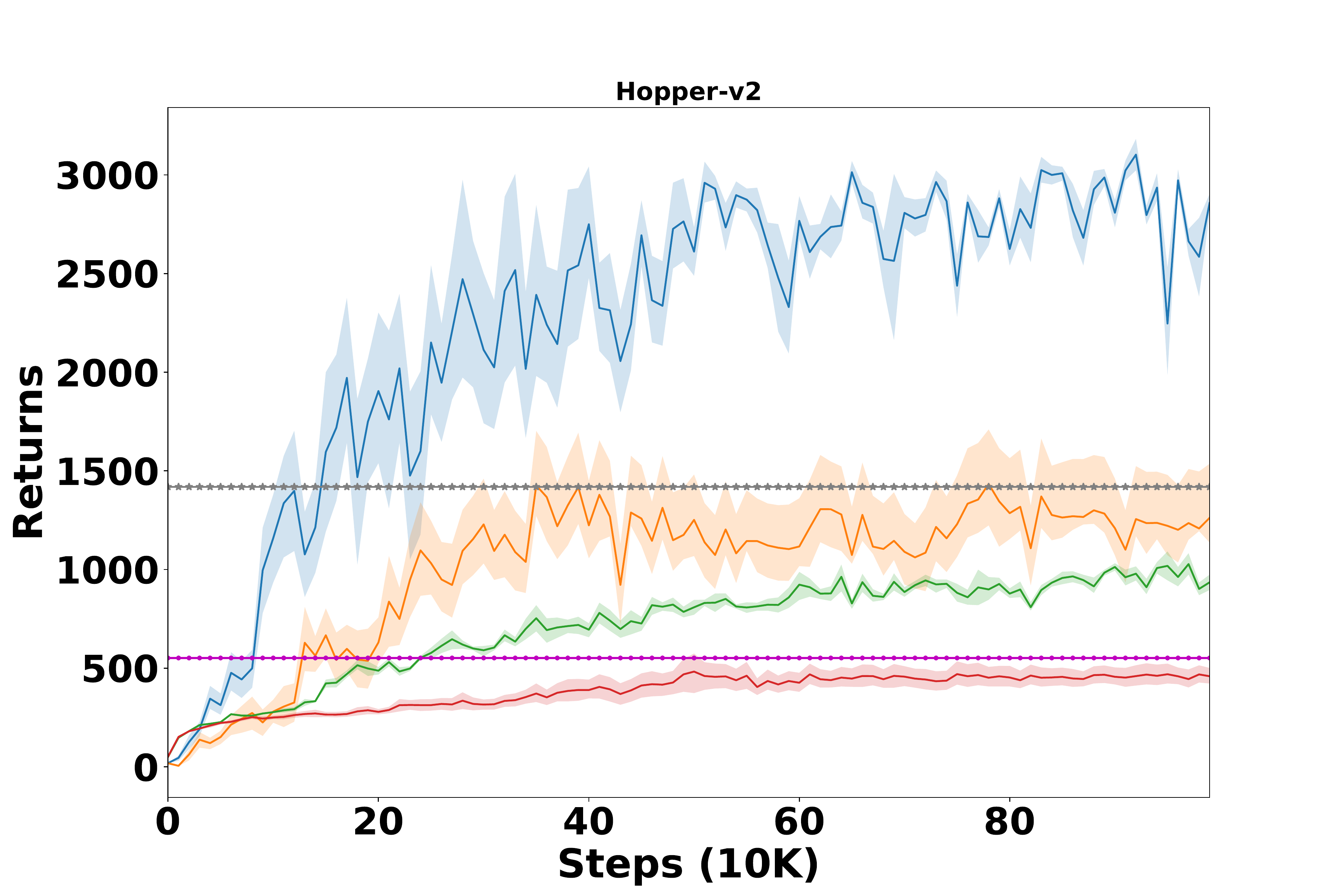}} 
        \end{minipage} 
        %\vskip -0.1in  
    \caption{Performance of \SAIL and other baselines using 4 suboptimal demonstration trajectories.}
    \label{fig:ep4-sail-dynamic}
    \end{center}
    %\vskip -0.2in
\end{figure*}

\textbf{Sample efficiency:}
As the results shown in Figure~\ref{fig:ep1-sail-dynamic}, 
\SAIL is the only method that performs consistently better in all 
tasks in terms of both sample efficiency and asymptotic performance. 
% Its sample efficiency are two-folds.
At the initial stage of the learning, \SAIL can quickly exploit 
the suboptimal demonstrations and approach to the demonstration's 
performance with significantly less samples. 
Furthermore, comparing Figure~\ref{fig:ep1-sail-dynamic} and Figure~\ref{fig:ep4-sail-dynamic}, we notice that \SAIL is highly 
effective in terms of utilizing the suboptimal demonstrations, as 
it is has negligible performance degradation across all tasks 
when we reduce the demonstration size from four trajectories to one. 
To see this, we highlight the performance comparison in task \textit{Swimmer-v2},
where GAIL and DAC reaches the demonstration performance 
with much more samples given 1 teacher trajectory, comparing to the case of 
four trajectories, whereas the performance of \SAIL is comparable in both cases.

% First, \SAIL is able to perform imitation learning using limited number of teacher demonstrations. 
%Figure \ref{fig:ep1-sail-dynamic} and Figure \ref{fig:ep4-sail-dynamic} shows the results when only 1 and 4 teacher demonstration trajectories are used for training, respectively.
% Figure \ref{fig:ep1-sail-dynamic} shows its performance when only 1 trajectory of teacher demonstration is provided for imitation learning.
% %
% Second, \SAIL is faster in terms of reaching the demonstration performance, with significantly less steps interacted with the environment.

\textbf{Effective exploration:}
Besides sample-efficiency, another advantage of 
\SAIL is that it can effectively explore the environment 
to achieves expert-level performance, even with highly sparse rewards. 
We observe that the prior solution of learning from environment rewards for exploration, such as POfD, cannot effectively address our proposed problem setting, as the delayed episodic feedback is too noisy to learn a meaningful critic.
Unlike other imitation learning baselines whose performance 
are limited by the demonstrations, \SAIL can rapidly surpass the imperfect 
teacher via constructing a better demonstration buffer and gradually
converge to near-optimal performance.
%Regular actor-critic approaches reply on environment rewards to learn critics, leading to high sample complexity in long-horizontal tasks with highly sparse rewards. 
%Contrasting to prior work, we leverage the sparse environment feedbacks only for selecting better trajectories than demonstration.
%A shaped reward function is built in place to ensure efficient imitation learning on those selected demonstrations.
%

\subsection{Analysis of \SAIL} \label{sec:abstudy}
In this section, we aim to analyze different components of our algorithm via ablation studies. Experiments in this section focus on the following aspects:

{\textbf{Effects of learning from expert demonstrations:}
As shown in Figure~\ref{fig:ep1-abstudy}, we observe that sampling from a mixture of teacher data and self-generated data accelerates the learning performance in early training stages.
Specifically, the blue line (\textit{SAIL-Dynamic}) refers our proposed approach. It initializes $\alpha$ to 0.5 and reduced $\alpha$ to zero once the learning policy is able to generate better-than-teacher trajectories. 
The orange line (\textit{SAIL}) represents a variant which adopts a fixed $\alpha=0.5$ throughout the training stages.
The green line (\textit{SAIL-without-LfD}) only uses self-generated data to learn policy by setting $\alpha=0$ constantly.
We see that the initial performance of \textit{SAIL-without-LfD} is less significant compared with its other two counterparts.
%
%\judycom{As we analyzed in Section \ref{sec:lfd-theory}, learning from demonstrations can be considered as a process of occupancy measure regularization.In early training stages, $\pi$ is close to a random policy whose performance is worse than the teacher. Setting a positive $\alpha$ regulates the divergence between $\dP$ and $\dE$ and can therefor warmup the imitation learning process. After $\pi$ is able to reproduce teacher-level trajectories, exploration plays a more important role, and adjusting $\alpha$ to zero brings the performance to (near-) optimal level across different tasks.}

\textbf{Effects of updating teacher demonstration buffers:} 
As shown in Figure~\ref{fig:ep1-abstudy}, the red line (\textit{SAIL-without-Expert-Adaptation}) refers to a variant of \SAIL which never update the teacher's replay buffer, even when a better trajectory is collected during the learning process.
We see that its asymptotic performance is bounded by the teacher's demonstration, which echoes the dilemma of most existing IL approaches. 
One key insight from these results is that, instead of learning critics based on sparse rewards from the environment, leveraging the sparse guidance to update the demonstration buffer, and then perform imitation learning on the updated demonstration, can be much more effective in improving the ultimate performance.

%simply adjusting the object we aim to imitate from is effective to improve 

\textbf{Benefits of exploration-driven objective:}
In order to illustrate the benefits of maximizing $\cE[\log(\frac{\dE}{\dB})]$ over an \IL objective such as $\cE[\log(\frac{\dE}{\dP})]$, we conduct a comparison study, where we train the discriminator using teacher demonstrations $\tau_T$ and on-policy self-generated samples $\tau_\pi$, instead of off-policy samples from replay buffer $\RB$. This on-policy training scheme is the same as proposed in \GAIL \cite{ho2016generative}. 
In this way, the discriminator can get approximations of $\log(\frac{\dE}{\dP})$ instead of $\log(\frac{\dE}{\dP})$.
We use the output of this on-policy discriminator to shape rewards, whereas $Q$ and $\pi$ are still updated in the same off-policy fashion as our proposed approach.
 
Performance of this approach is illustrated by the orange line (\textit{SAIL-OnPolicy}) in Figure \ref{fig:ep4-onpolicy-abstudy}. Compared to the on-policy \GAIL whose performance is illustrated by the green line, \textit{SAIL-OnPolicy} still enjoys the benefits of off-policy actor-critic learning scheme in general. However, it is less effective compared with our proposed approach. Even when $\pi$ and $Q$ are updated using off-policy, \textit{SAIL-OnPolicy} is obviously slower to surpass the teacher demonstration (dashed gray line), due to its pure imitation-driven objective.
This results verify that our objective is more effective in terms of encouraging exploration.
\SAIL enjoys fast improvement in performance not only because of an adaptive teacher demonstration buffer, but also because it is guided by exploration-driven shaped reward functions.

\begin{figure*}[ht]
    %\vskip 0.2in
    \begin{center}
        %%%%%%%%%%%%%%%%%%%%%%%%%%%%%%%%%%%%%%%%%%%%%%%%%%%%%%%%%%%%%%%%%%%%
        \hskip -0.1in 
        \begin{minipage}[b]{.27\textwidth}
            \centerline{\includegraphics[width=\columnwidth]{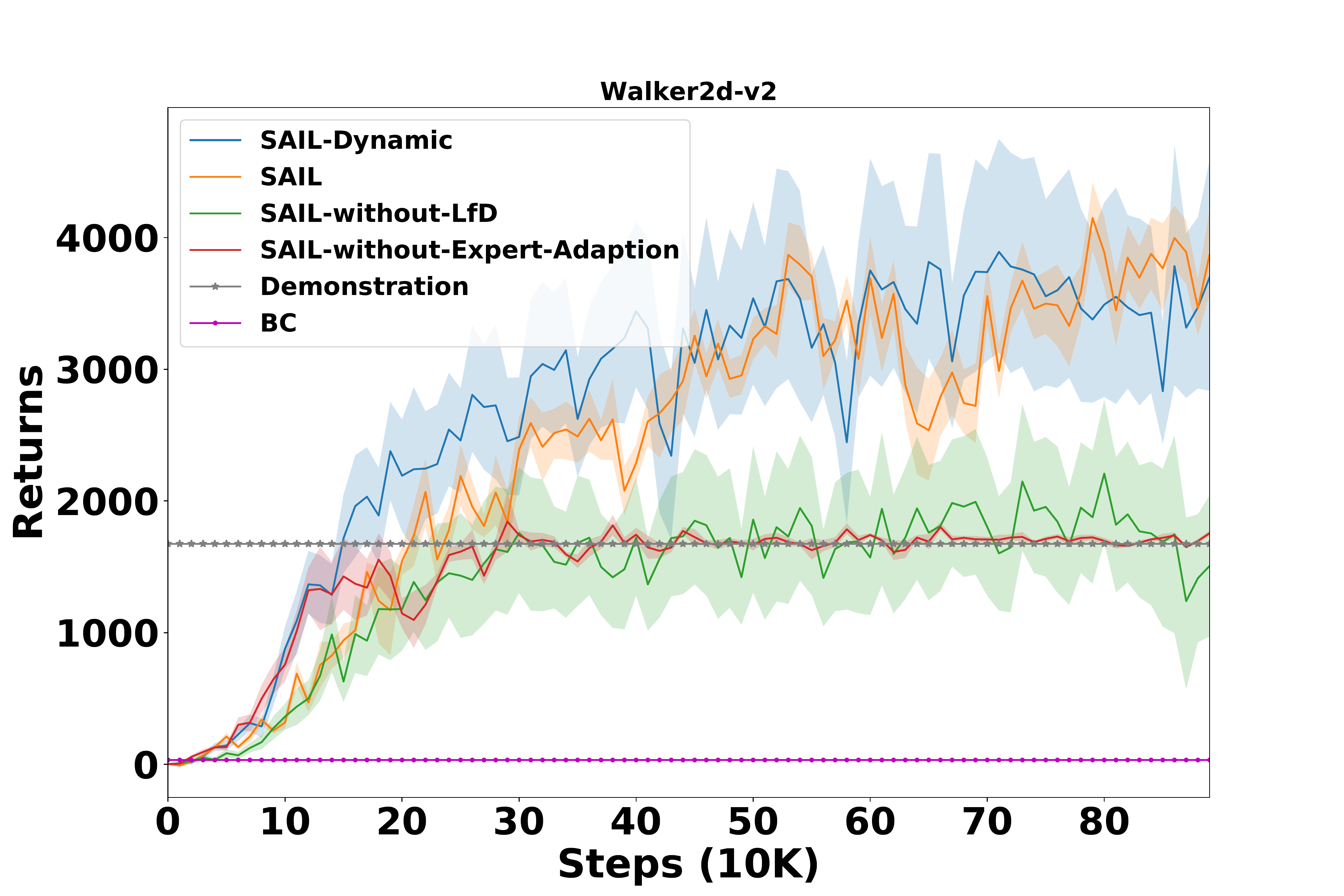}} 
        \end{minipage} \hskip -0.2in %\qquad  
        \begin{minipage}[b]{.27\textwidth}
            \centerline{\includegraphics[width=\columnwidth]{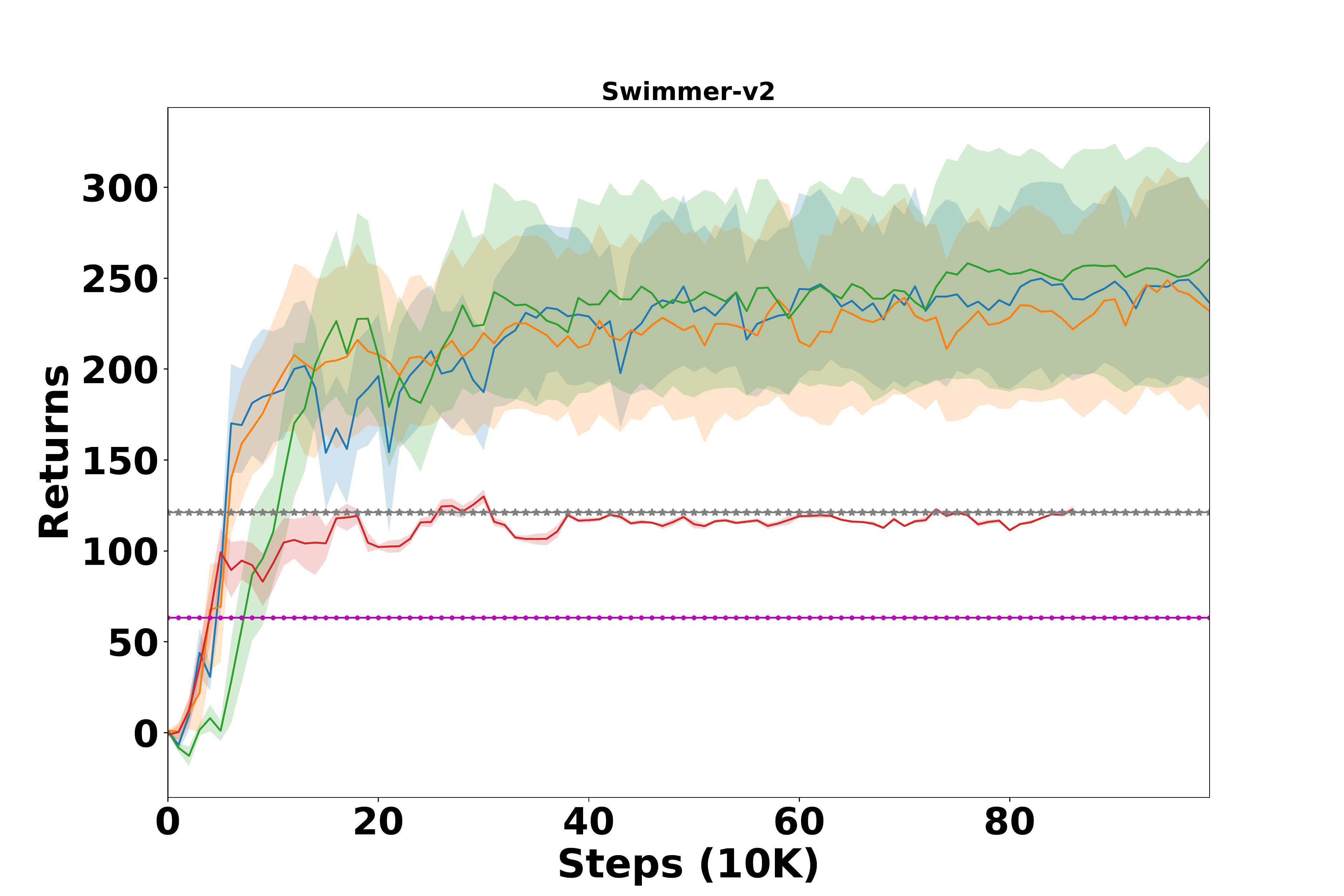}}
            %\caption{Swimmer-v2}\label{ep4-swimmer}
        \end{minipage} \hskip -0.2in 
        \begin{minipage}[b]{.27\textwidth}
            \centerline{\includegraphics[width=\columnwidth]{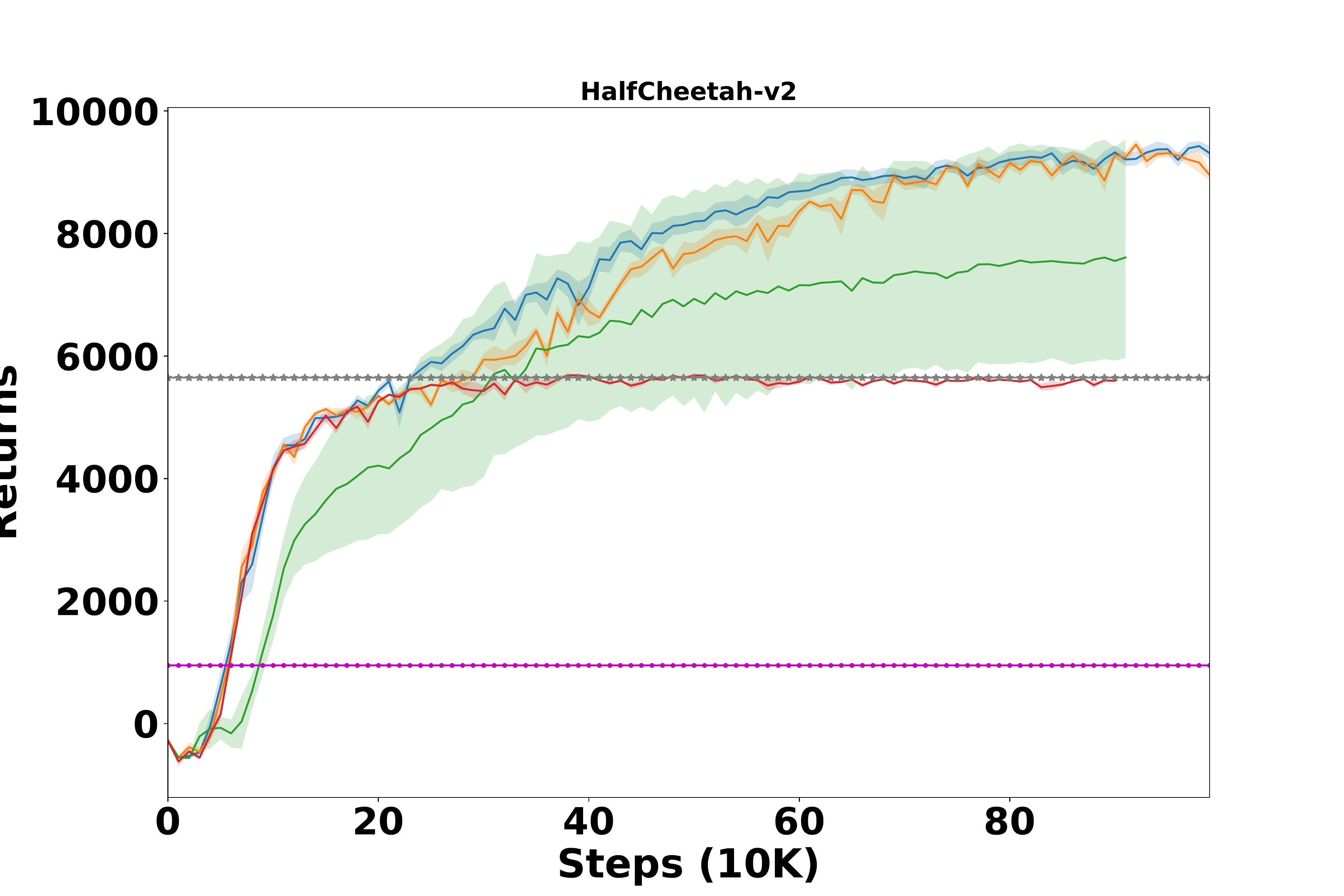}} 
        \end{minipage} \hskip -0.2in   
        \begin{minipage}[b]{.27\textwidth}
            \centerline{\includegraphics[width=\columnwidth]{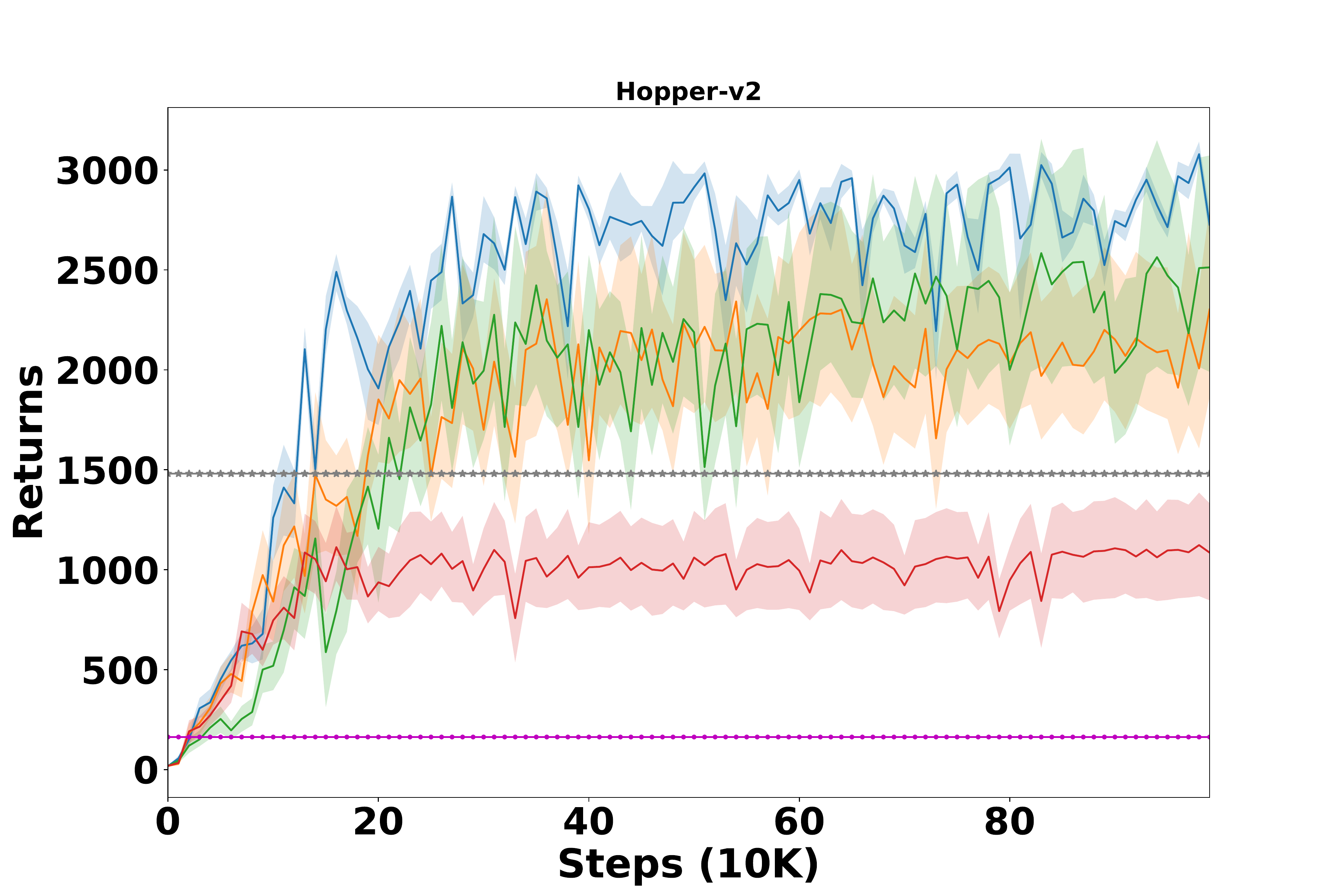}} 
        \end{minipage}   
        %\vskip -0.1in  
        \caption{Ablation Study of \SAIL using 1 teacher demonstration trajectory.} \label{fig:ep1-abstudy} 
        %%%%%%%%%%%%%%%%%%%%%%%%%%%%%%%%%%%%%%%%%%%%%%%%%%%%%%%%%%%%%%%%%%%% 
        \hskip -0.1in  
        \begin{minipage}[b]{.27\textwidth}
            \centerline{\includegraphics[width=\columnwidth]{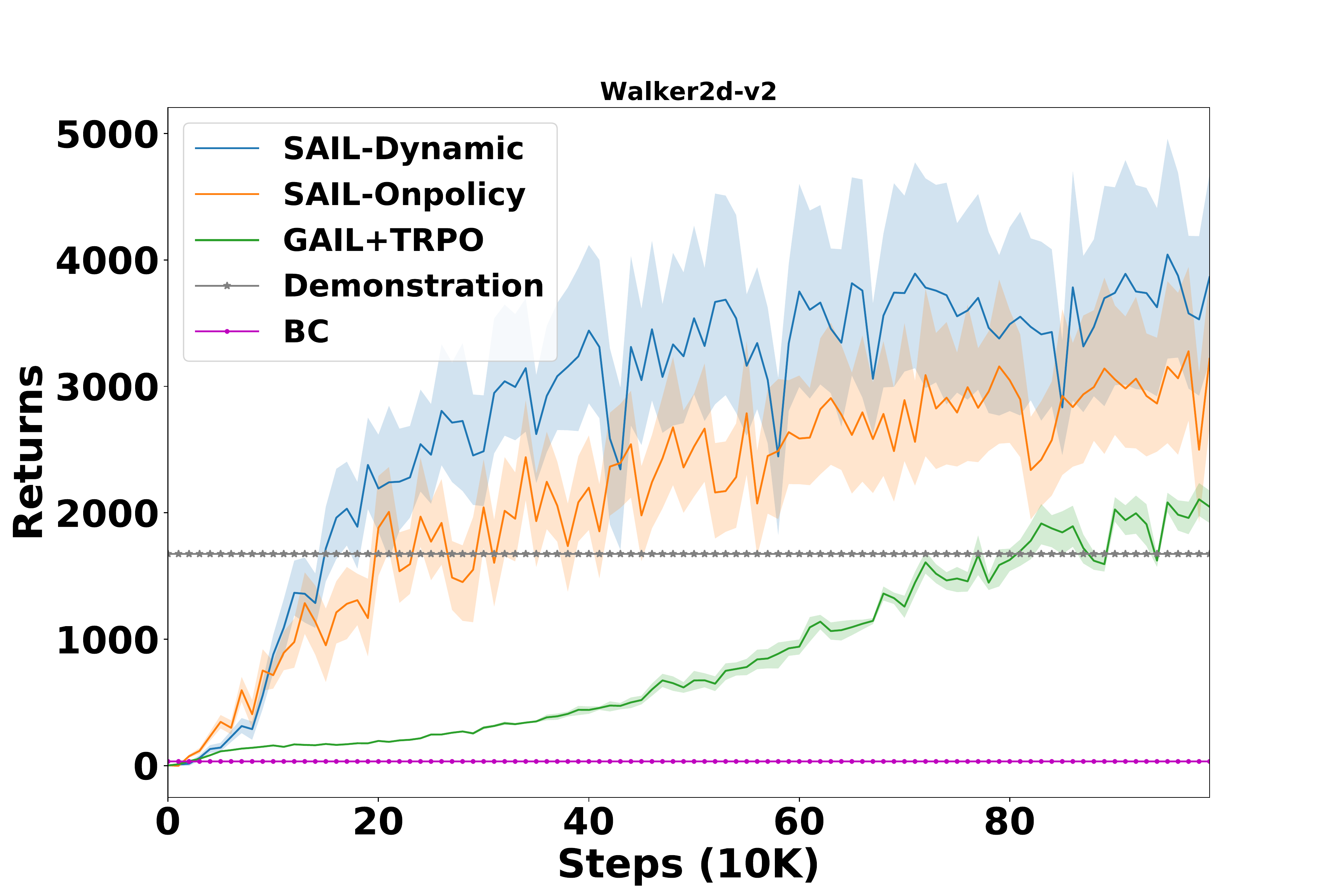}} 
        \end{minipage}  \hskip -0.2in  
        \begin{minipage}[b]{.27\textwidth}
            \centerline{\includegraphics[width=\columnwidth]{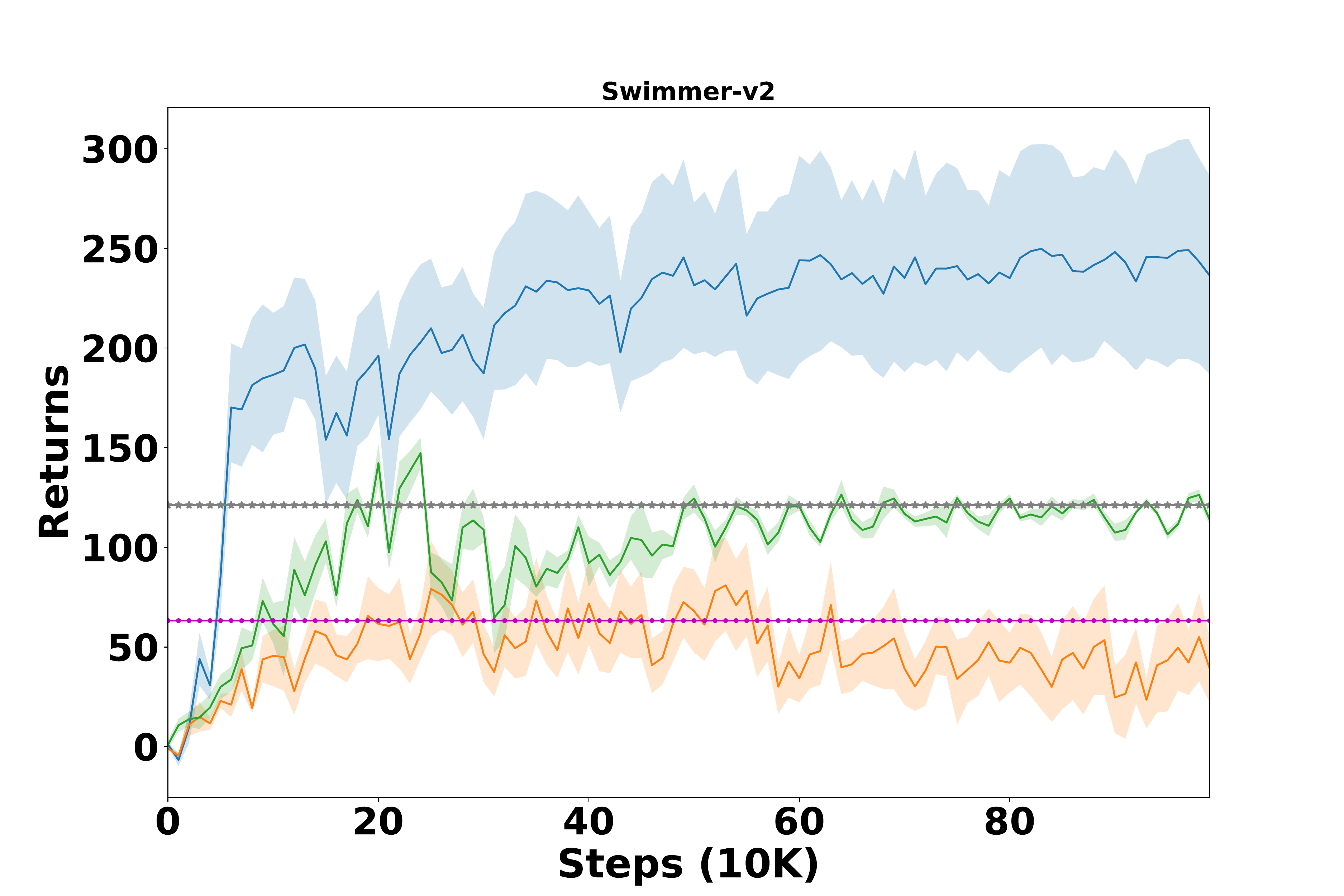}} 
        \end{minipage} \hskip -0.2in 
        \begin{minipage}[b]{.27\textwidth}
            \centerline{\includegraphics[width=\columnwidth]{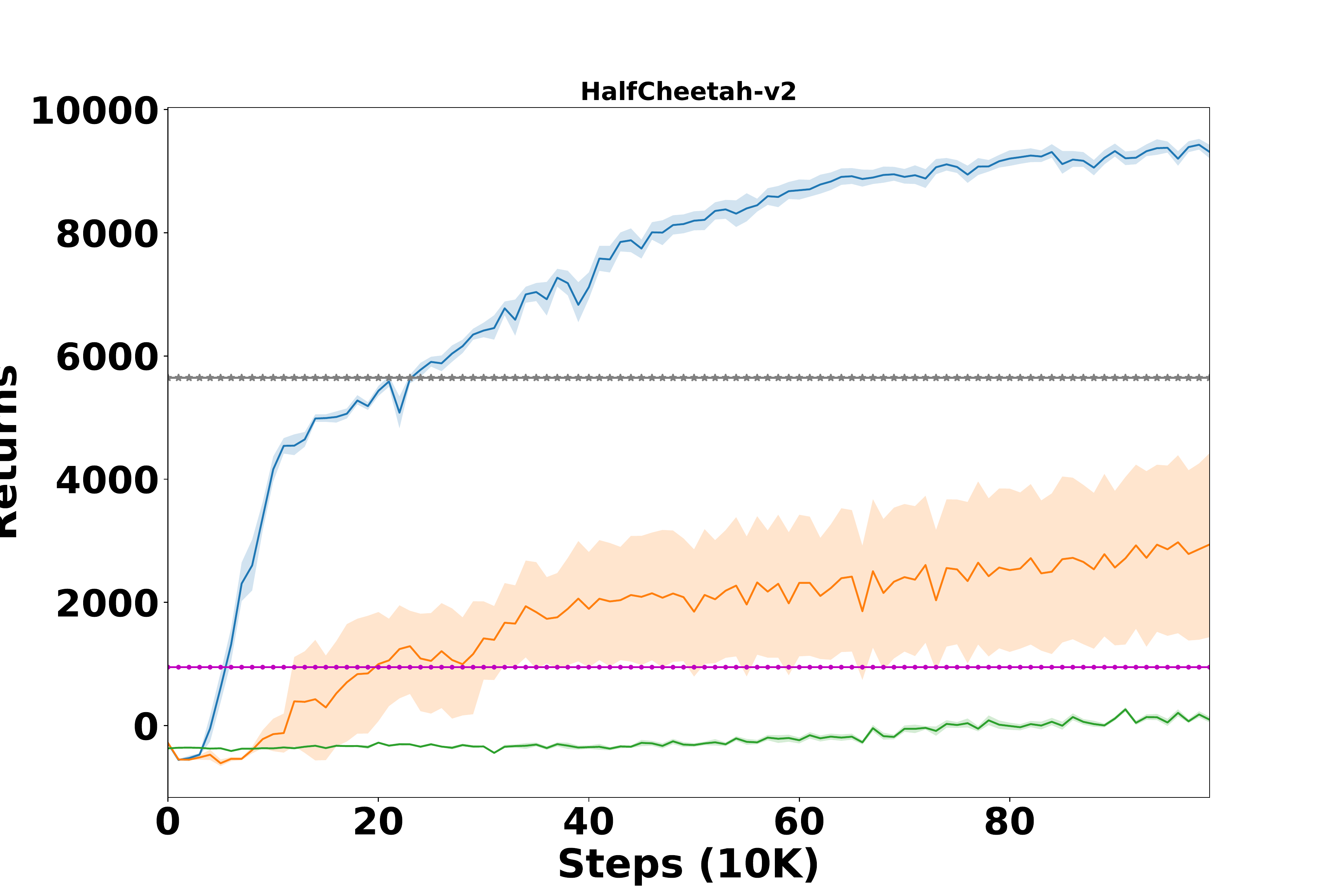}}
            %\caption{Swimmer-v2}\label{ep4-swimmer}
        \end{minipage} \hskip -0.2in  
        \begin{minipage}[b]{.27\textwidth}
            \centerline{\includegraphics[width=\columnwidth]{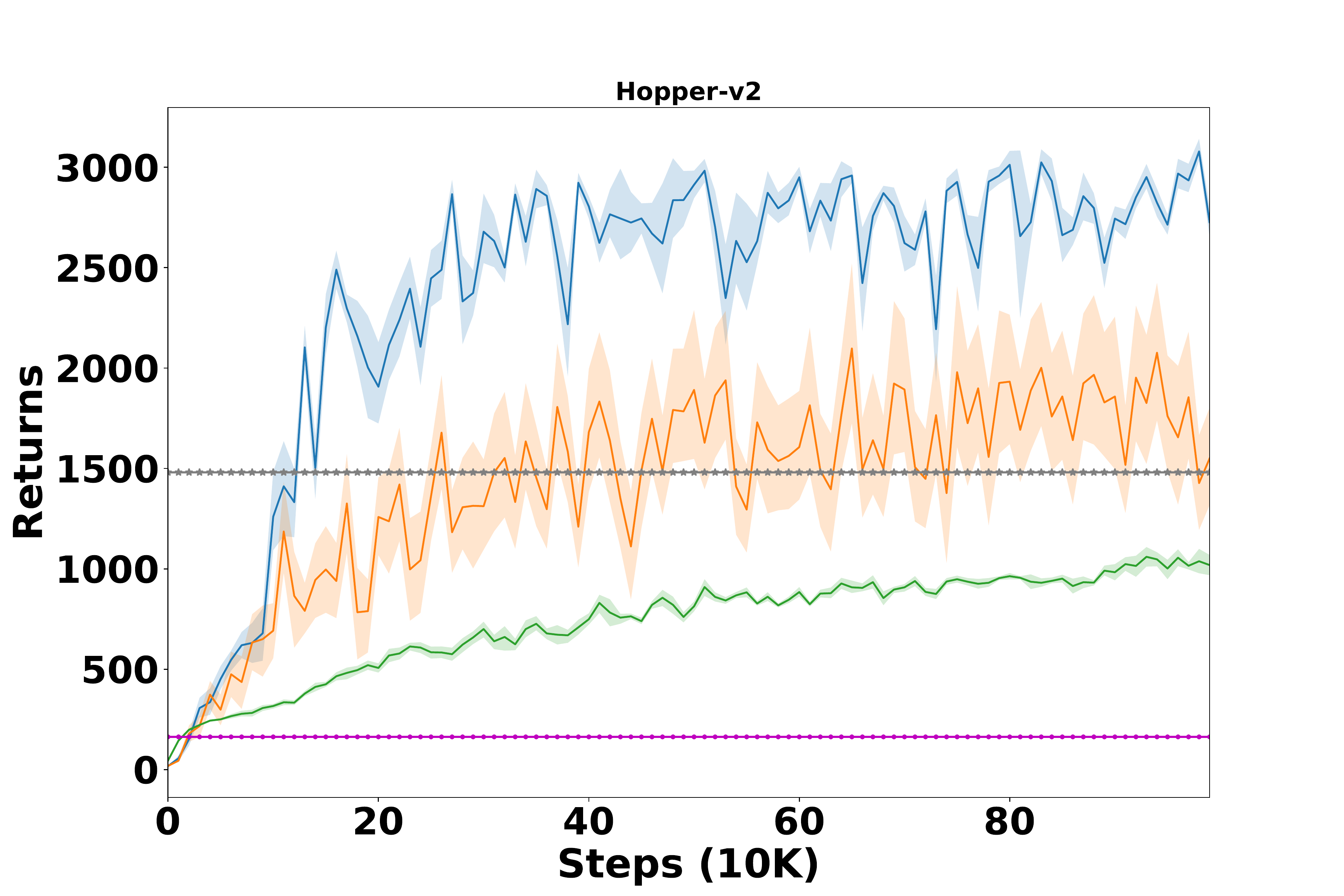}}
            %\caption{Swimmer-v2}\label{ep4-swimmer}
        \end{minipage} %\hskip -0.2in  
        %\vskip -0.1in  
        \caption{Comparison of off-policy and on-policy reward shaping using 1 teacher demonstration trajectory.} \label{fig:ab-ep4} 
    \label{fig:ep4-onpolicy-abstudy}
    \end{center}
    %\vskip -0.2in
\end{figure*}

% \subsection{Sensitivity of SAIL to expert demonstration buffer size} 

% %\judycom{demonstration buffer size: 5e3, 1e4, 5e4, 1e5, 5e5}

% A key component of our algorithm is adjusting the expert demonstration buffer by adding new trajectories with high episodic rewards. There is a tradeoff between the number of trajectory we keep in the buffer and the robustness of our algorithm.

% \subsection{Sensitivity of SAIL to self-generated buffer size}
% \judycom{Focus on games like Swimmer-v2, Reacher-v2 on which BC performs perfectly.}

% Reacher-v2 now: has 1e6 buffer size
% !TEX ROOT=../../main.tex

\section{Related Work}
{\textbf{Imitation Learning (IL)}} is originally derived from Behavior Cloning (BC), which maximizes the probability for a policy to follow the demonstrated actions by conducting supervise learning on the demonstration data~\cite{schaal1999imitation}. BC is known to suffer from the out-of-distribution errors due to the shift-propagation issue.
Later successes in {Inverse Reinforcement Learning (IRL)} address IL from a different perspective~\cite{abbeel2004apprenticeship,ziebart2008maximum}. IRL replicates the expertise by first recovering a reward function that best explains the expert behavior, then performing RL based on the learned reward function. 
An instantiation of the IRL approach is \cite{syed2008game}, which treats imitation learning as a two-player game to recover the reward and policy interactively.

Later work of IRL draws a connection to Generative Adversarial Training~\cite{goodfellow2014generative} to solve imitation learning as a distribution matching problem.
Especially, \cite{ho2016generative} proposed Generative Adversarial Imitation Learning ({GAIL}) to learn the expert policy without needing to unveil its reward functions.
%A similar work is {AIRL} \cite{fu2017learning} which further extended GAIL to be able to potentially recover the reward function.
%
The principle of GAIL has been applied to different on-policy RL frameworks, including PPO~\cite{kang2018policy}, TRPO~\cite{wu2019imitation,fu2017learning}, etc.
These on-policy approaches require a large number of interactions with the environment. 

Recently, there are IL approaches that address the distribution matching problem under off-policy actor-critic (OPAC) frameworks. 
\cite{kostrikov2018discriminator} proposed Discriminator Actor Critic (DAC), which applies the methodology of GAIL on TD3 framework.
They propose to learn a discriminator with off-policy samples and correct the distribution shift by importance sampling. %, which was actually omitted in their implementation.
To further stabilize the performance of IL, they altered benchmark environments by attaching an extra \textit{absorbing state} to the end of each episode.
Other compatible work about off-policy IL can also be found in \cite{sasaki2018sample,kostrikov2019imitation}, which shares a common objective to minimize $D[\dP||d_E]$, \ie~the density divergence between the teacher and learning policy.
%who defined a shaped reward function to be the probability that state action comes from an expert policy, so that a trajectory-wise rewards can be considered as accumulated log-likelihood of a Bernoulli distribution. 
%Another work along the line of off-policy imitation learning is ValueDICE \cite{kostrikov2019imitation}, which is extended from DualDICE \cite{nachum2019dualdice} that estimates the stationary distribution corrections using off-line data.

Some other work address IL by reducing distribution shift instead of directly seeking distribution matching. \cite{ross2011reduction} proposed \textit{DAGGER} algorithm to obtain linear regret on imitation learning, provided that they have feedback from an oracle policy.
\cite{burda2019exploration} applied the Random Network Distillation \cite{wang2019random} to leverage the curiosity loss from a random network as penalty, forcing the density distribution of the learned policy to stay close to the expert demonstrations.
\cite{brantleydisagreement} derived IL approaches via policy ensemble: they pre-trained multiple policies by BC, and then used the variance among these pre-trained policies as penalty to regularize the target policy. 

Contrasting to our exploration-driven objective, prior arts mentioned above are motivated to exactly recover the teacher policy.
% neglects the potential of utilizing self-explored trajectories to improve the upper-bound of the imitation target
Addressing a different and more practical problem setting, our work is able to significantly surpass the teacher, by iteratively playing a tradeoff between imitation and exploration.
 
{\textbf{Learning from Demonstrations (LfD)}} address the challenge of exploration  faced by model-free RL by leveraging expert demonstration as self-generated transitions.  
Especially, \cite{hester2018deep} introduced Deep $Q$-learning from Demonstrations (DQfD) for tasks with discrete action spaces. 
\cite{vevcerik2017leveraging} applied the principle of LfD to the DDPG framework \cite{lillicrap2015continuous} to facilitate learning in continuous action spaces.
%Prior work along this line assume that the demonstrations are either sufficient in quantity or near-optimal in performance.
%
Strategies such as prioritized sampling \cite{hester2018deep} or refined cost functions \cite{kim2013learning,vevcerik2017leveraging} are usually required for LfD approaches to encourage following the expert actions.
We inherit the same spirit of LfD in that we combine teacher demonstration with self-generated data to accelerate learning in early stages. However, we omit the need of a prioritized scheme or hand-crafted loss function, benefiting from a shaped reward function which naturally ensures the superiority of the teacher demonstration over self-generated data.

Recently, there emerges research on learning from sub-optimal demonstrations~\cite{sun2018truncated,kang2018policy,wu2019imitation}.
To surpass the performance of the demonstrations, they alter the environment reward to combine it with an auxiliary term, based on which a policy is learned to optimize the accumulated returns.
Contrasting to their work, our approach learns a critic without using intermediate environment rewards, which is a more robust learning scheme when the environment feedbacks are highly sparse or delayed. 
\SAIL learns a policy based on synthetic rewards from a discriminator, which does not reveal the true reward signals but purely represents the occupancy match between the self-generated data (for exploration) and given demonstrations (for imitation).
%Instead of relying on sparse environment rewards to update policy, we leverage them only to select self-generated data to update the teacher demonstrations buffer, in order to improve the lower-bound of .

%
%requires training data with dense environment rewards to fully utilize the expert demonstration. Moreover, they 
\section{Conclusion}
In this paper, we address the problem of reinforcement learning in environments with highly \textit{sparse} rewards given \textit{sub-optimal} teacher demonstrations.
To address this challenging problem, we propose an novel objective which encourages exploration-based imitation learning. Towards this objective we design an effective algorithm called \textit{Self Adaptive Imitation Learning (SAIL)}.
\SAIL is validated to 1) address sample efficiency by off-policy imitation learning, 2) accelerate learning process by fully utilizing teacher demonstration, and 3) surpass the imperfect teacher to reach (near) optimality by iteratively performing imitation and exploration.
Experimental results across modified locomotion tasks with highly sparse rewards indicate that, \SAIL significantly surpasses state-of-the-arts in terms of both sample efficiency and asymptotic performance.

\bibliography{ref}
\bibliographystyle{unsrt}

\end{document}